\documentclass[runningheads]{llncs}
\usepackage{graphicx}
\usepackage{tikz}
\usepackage{comment}
\usepackage{amsmath,amssymb} % define this before the line numbering.
\usepackage{color}

\usepackage[accsupp]{axessibility}  % Improves PDF readability for those with disabilities.
\usepackage[nocompress]{cite}
\usepackage{xcolor}
\usepackage{nicefrac}
\usepackage{booktabs}
\usepackage{multirow}
\usepackage{algpseudocode}
\usepackage{algorithm}
\usepackage[bb=boondox]{mathalfa}
\usepackage{caption}
\usepackage{subcaption}

\begin{document}

\pagestyle{headings}
\mainmatter

\title{Diffusion Models for Counterfactual Explanations} 

% CAMERA READY SUBMISSION
\titlerunning{Diffusion Models for Counterfactual Explanations}

\author{Guillaume Jeanneret, Lo\"{i}c Simon and Frédéric Jurie}
\authorrunning{G. Jeanneret et al.}

\institute{Normandy University, ENSICAEN, UNICAEN, CNRS, GREYC, France \\
{\tt\small guillaume.jeanneret-sanmiguel@unicaen.fr}}

\maketitle
\begin{abstract}
Counterfactual explanations have shown promising results as a post-hoc framework to make image classifiers more explainable. In this paper, we propose DiME, a method allowing the generation of counterfactual images using the recent diffusion models. By leveraging the guided generative diffusion process, our proposed methodology shows how to use the gradients of the target classifier to generate counterfactual explanations of input instances. Further, we analyze current approaches to evaluate spurious correlations and extend the evaluation measurements by proposing a new metric: Correlation Difference. Our experimental validations show that the proposed algorithm surpasses previous State-of-the-Art results on 5 out of 6 metrics on CelebA.
\end{abstract}
\section{Introduction}

Convolutional neural networks (CNNs) reached performances unimaginable a few decades ago, thanks to the adoption of very large and deep models (e.g. with hundreds of layers and nearly billions of trainable parameters). Yet, it is difficult to explain their decisions because they are highly non-linear and over-parametrized. Moreover, for real-life applications, if a model exploits spurious correlations of data to forecast a prediction, the end-user will doubt the validity of the decision. Particularly, in high-stake scenarios like medicine or critical systems, ML must guarantee the usage of correct features to compute a prediction and prevent counterfeit associations. For this reason, the Explainable Artificial Intelligence (XAI) research field has been growing in recent years to progress towards understanding the decision-making mechanisms in black-box models. 

In this paper, we focus on \emph{post-hoc} explanation methods. Notably, we concentrate on the growing branch of Counterfactual Explanations (CE)~\cite{wachter2018CounterfactualExplanationsOpening}. CEs aim to create minimal but meaningful perturbations of an input sample to change the original decision given by a black-box model. 
Although the objective between CE and adversarial examples share some similarities~\cite{pawelczykExploringCounterfactualExplanations2021}, the CEs' perturbations must be understandable. In contrast, adversarial examples~\cite{madry2018towards} contain high-frequency noise \textit{indistinguishable} for the human eye.
On overall, CEs target three goals: \textit{(i)} create proximal images with sparse modifications, \textit{i.e.} instances with the smallest perturbation, \textit{(ii)} the explanations must be realistic and understandable by a human, and \textit{(iii)} the counterfactual generation method must create diverse instances. In general, counterfactual explanations seek to reveal the learned correlations related to the model's decisions.

Multiple works on CE use generative models to create tangible changes in the image~\cite{sauer2021counterfactual,Rodriguez_2021_ICCV,Joshi2018xGEMsGE}. Further, these architectures recognize the factors to generate images near the image-manifold~\cite{arora2018do}. Given the recent advances within image synthesis community, we propose DiME: \underline{Di}ffusion \underline{M}odels for counterfactual \underline{E}xplanations. DiME harnesses the denoising diffusion probabilistic models~\cite{NEURIPS2020_4c5bcfec} to produce CEs. For simplicity, we will refer to these models as diffusion models or DDPMs. To the best of our knowledge, we are the first to exploit these new synthesis methods in the context of CE. 

Diffusion models offer several advantages compared to alternate generative models, such as GANs. First of all, DDPMs have several latent spaces; each one controls coarse and fine-grained details. We take advantage of low-level noise latent spaces to generate semantically-meaningfully changes in the input image. These spaces only have been recently studied by~\cite{meng2022sdedit} for inpainting. Secondly, due to their probabilistic nature, they produce a diverse set of images. Stochasticity is ideal for CEs because multiple explanations may explain a classifier's error modes. Third, Nichol and Dhariwal~\cite{nichol2021improved} results suggest that DDPMs cover a broader range of the target image distribution. Indeed, they noticed that for similar FID, the recall is much higher on the improved precision-recall metrics~\cite{Kynkaanniemi2019}. Finally, DDPMs' training is more stable than the State-of-the-Art synthesis models, notably GANs. Due to their relatively new development, DDPMs are under-studied, and multiple aspects are yet to be deciphered. 

We contribute a small step into the XAI community by studying the low-level noised latent spaces of DDPMs in the context of counterfactual explanations. We summarize our contributions as follows:
\begin{itemize}
    \item DiME uses the recent diffusion models to generate counterfactual examples. Unlike other generative models, our CE algorithm does not require training the diffusion model in a conditioned way or retraining it using gradients, \emph{i.e.} we rely on a single trained unconditional DDPM to achieve our objective.
    \item We derive a new way to leverage an existing (target) classifier to guide the generation process instead of ones trained on noisy instances. 
    \item We set a new State-of-the-Art result on CelebA, surpassing the previous works on counterfactual explanations on the FID, FVA, and MNAC metrics for the \textit{Smile} attribute and the FID and MNAC for the \textit{Young} feature.
    \item We show that the MNAC provides a false sense of evaluating counterfactuals correctly. So we introduce a new metric, dubbed Correlation Difference, to evaluate subtle spurious correlations on a CE setting.
\end{itemize}

\section{Related Work}

Our work contributes to the field of XAI, within which two families can be distinguished: interpretable-by-design and \emph{post-hoc} approaches. The former includes, at the design stage, human interpretable mechanisms~\cite{alvarezmelis2018towards,Zhang_2018_CVPR,NEURIPS2019_adf7ee2d,Nauta_2021_CVPR,Alaniz_2021_CVPR,Huang_2020_CVPR,Bohle_2021_CVPR}. The latter aims at understanding the behavior of existing ML models without modifying their internal structure. Our method belongs within this second family. The two have different objectives and advantages; one benefit of \emph{post-hoc} methods is that they rely on existing models that are known to have good performance, whereas XAI by design often leads to a performance trade-off.

\textbf{Post-hoc methods:} In the field of \emph{post-hoc} methods, there are several explored directions. Model Distillation strategies~\cite{tan2018learning,Ge_2021_CVPR} approach explainability through fitting an interpretable model on the black-box models' predictions. In a different vein, some methods generate explanation in textual form~\cite{10.1007/978-3-319-46493-0_1,Park_2018_CVPR,Xian_2019_CVPR}. When it comes to explaining visual information, feature importance is arguably the most common approach, often implemented in the form of saliency maps computed either using the gradients within the network~\cite{Selvaraju_2017_ICCV,Jalwana_2021_CVPR,Wang_2020_CVPR_Workshops,Lee_2021_CVPR,8354201,Zhou_2016_CVPR} or using the perturbations on the image~\cite{DBLP:conf/bmvc/PetsiukDS18,DBLP:conf/cvpr/PetsiukJMM0OS21,Vasu_2020_WACV,Hatakeyama_2020_ACCV}. Concept attribution methods seek the most recurrent traits that describe a particular class or instance. Intuitively, concept attribution algorithms use~\cite{pmlr-v80-kim18d} or search~\cite{NEURIPS2019_77d2afcb,Yeh2020OnCC,Ge_2021_CVPR,Zhou_2018_ECCV} for human-interpretable notions such as textures or shapes. 

\textbf{Counterfactual Explanations (CEs):} CEs is a branch of post-hoc explanations. They are relevant to legally justify decisions made automatically by algorithms \cite{wachter2018CounterfactualExplanationsOpening}. In a nutshell, a CE is the smallest meaningful change to an input sample to obtain a desirable outcome of the algorithm. Some recent methods~\cite{Wang_2021_CVPR,DBLP:conf/icml/GoyalWEBPL19} exploit the query image's regions and a different classified picture to interchange semantic appearances, creating counterfactual examples. Other works~\cite{wachter2018CounterfactualExplanationsOpening,pmlr-v130-schut21a} leverage the input image's gradients with respect to the target label to create meaningful perturbations. Conversely, \cite{Akula_Wang_Zhu_2020} find patterns via prototypes that the image must contain to alter its prediction. Similarly, \cite{Poyiadzi2020FACEFA,looveren2021interpretable} follow a prototype-based algorithm to generate the explanations. Even Deep Image Priors~\cite{thiagarajan2021designing} and Invertible CNNs~\cite{hvilshoj2021ecinn} have shown the capacity to produce counterfactual examples. Furthermore, theoretical analyses~\cite{NEURIPS2019_7392ea4c} found similarities between counterfactual explanations and adversarial attacks.

Due to the nature of the problem, the generation technique used is the key element to produce data near the image manifold. For instance, \cite{NEURIPS2018_c5ff2543} optimizes the residual of the image directly using an autoencoder as a regularizer. Other works propose to use generative networks to create the CEs, either unconditional~\cite{nemirovsky2020countergan,Rodriguez_2021_ICCV, shih2021GANMEXOnevsoneAttributions, zhao2018GeneratingNaturalAdversarial} or conditional~\cite{van2021conditional,Singla2020Explanation,8969491}. In this paper, we adopt more recent generation approaches, namely {\em diffusion models}; an attempt never considered in the past for counterfactual generation.

\textbf{Diffusion Models:} Diffusion models have recently gained popularity in the image generation research field~\cite{NEURIPS2020_4c5bcfec,song2021denoising}. For instance, DDPMs approached inpainting~\cite{saharia2021palette}, conditional and unconditional image synthesis~\cite{nichol2021improved,NEURIPS2020_4c5bcfec,Choi_2021_ICCV}, super-resolution~\cite{Saharia2021ImageSV}, even fundamental tasks such as segmentation~\cite{baranchuk2022labelefficient}, providing performance similar or even better than State-of-the-Art generative models. Further, studies like~\cite{song2021scorebased,huang2021a} show score-based approaches and diffusion are alternative formulations to denoise the reverse sampling for data generation. Due to the recursive generation process, DDPMs sampling is expensive. Many works have studied alternative approaches to accelerate the generation process~\cite{kong2021on,DBLP:journals/corr/abs-2106-03802}. 

The recent method of \cite{Dhariwal2021DiffusionMB} targets conditional image generation with diffusion models, which they do by training a specific classifier on noisy instances to bias the generation process. Our work bears some similarities to this method, but, in our case, explaining an existing classifier trained uniquely in clean instances poses an additional challenge. In addition, unlike past diffusion methods, we perform the image editing process from an intermediate step rather than the final one. To the best of our knowledge, no former study has considered diffusion models to explain a neural network counterfactually.
\section{Methodology}

\subsection{Preliminaries}

We begin by introducing the generation process of diffusion models. They rely on two Markov chain sampling schemes that are inverse of one another. In the forward direction, the sampling starts from a natural image $x$ and iteratively sample $z_1,\cdots,z_T$ by replacing part of the signal with white Gaussian noise. More precisely, letting $\beta_t$ be a prescribed variance, the forward process follows the recursive expression:
\begin{equation} \label{eq:degradation_onestep}
    z_t \sim \mathcal{N} (\sqrt{1-\beta_t}\,z_{t-1},  \beta_t \,I),
\end{equation}
where $\mathcal{N}$ is the normal distribution, $I$ the identity matrix, and $z_0 = x$. In fact, this process can be simulated directly from the original sample with 
\begin{equation} \label{eq:degradation}
    z_t \sim \mathcal{N} (\sqrt{\alpha_t}x, (1 - \alpha_t) I),
\end{equation}
where $\alpha_t := \prod_{k=1}^t (1 - \beta_k)$. For clarification, through the rest of the paper, we will refer to clean images with an $x$, while noisy ones with a $z$.

In the reverse process, a neural network recurrently denoises $z_T$ to recover the previous samples $z_{T-1},\cdots, z_0$. This network takes the current time step $t$ and a noisy sample $z_t$ as inputs, and produces an average sample $\mu(t, z_t)$ and a covariance matrix $\Sigma(t,z_t)$, shorthanded as $\mu(z_t)$ and $\Sigma(z_t)$, respectively. Then $z_{t-1}$ is sampled with 
\begin{equation} \label{eq:sampling}
    z_{t-1} \sim \mathcal{N}(\mu(z_t), \Sigma(z_t)).
\end{equation}
So, the DDPM algorithm iteratively employs Eq.~\ref{eq:sampling} to generate an image $z_0$ with zero variance, \textit{i.e.} a clean image. Some diffusion models use external information, such as labels, to condition the denoising process. However, in this paper, we employ an unconditional DDPM. 

In practice, the series of variances $\{\beta_t\}$ are chosen such that $z_T\sim \mathcal{N}(0,I)$. Further, the DDPM's trainable parameters are fitted so that the reverse and forward processes share the same distribution. For details on training schemes, we recommend the studies of Ho \emph{et al.}~\cite{NEURIPS2020_4c5bcfec} and Nichol and Dhariwal~\cite{nichol2021improved} to the reader. Once the network is trained, one can rely on the reverse Markov chain process to generate a clean image from a random noise image $z_T$. Besides, the sampling procedure can be adapted to optimize some properties following the so-called \emph{guided diffusion} scheme proposed in \cite{Dhariwal2021DiffusionMB}\footnote{In \cite{Dhariwal2021DiffusionMB}, the guided diffusion is restricted to a specific classification loss. Still, for the sake of generality and conciseness, we provide its extension to an arbitrary loss}:
\begin{equation} \label{eq:generalized-diff}
    z_{t-1} \sim \mathcal{N}(\mu(z_t) - \Sigma(z_t)\, \nabla_{z_t} L(z_t; y), \Sigma(z_t)),
\end{equation}
where $L$ is a loss function using $z_t$ to specify the wanted property of the generated image, for example, to condition the generation on a prescribed label $y$.

\subsection{DiME: Diffusion Models for Counterfactual Explanations}

\begin{figure*}[t]
    \centering
    \includegraphics[width=0.98\textwidth]{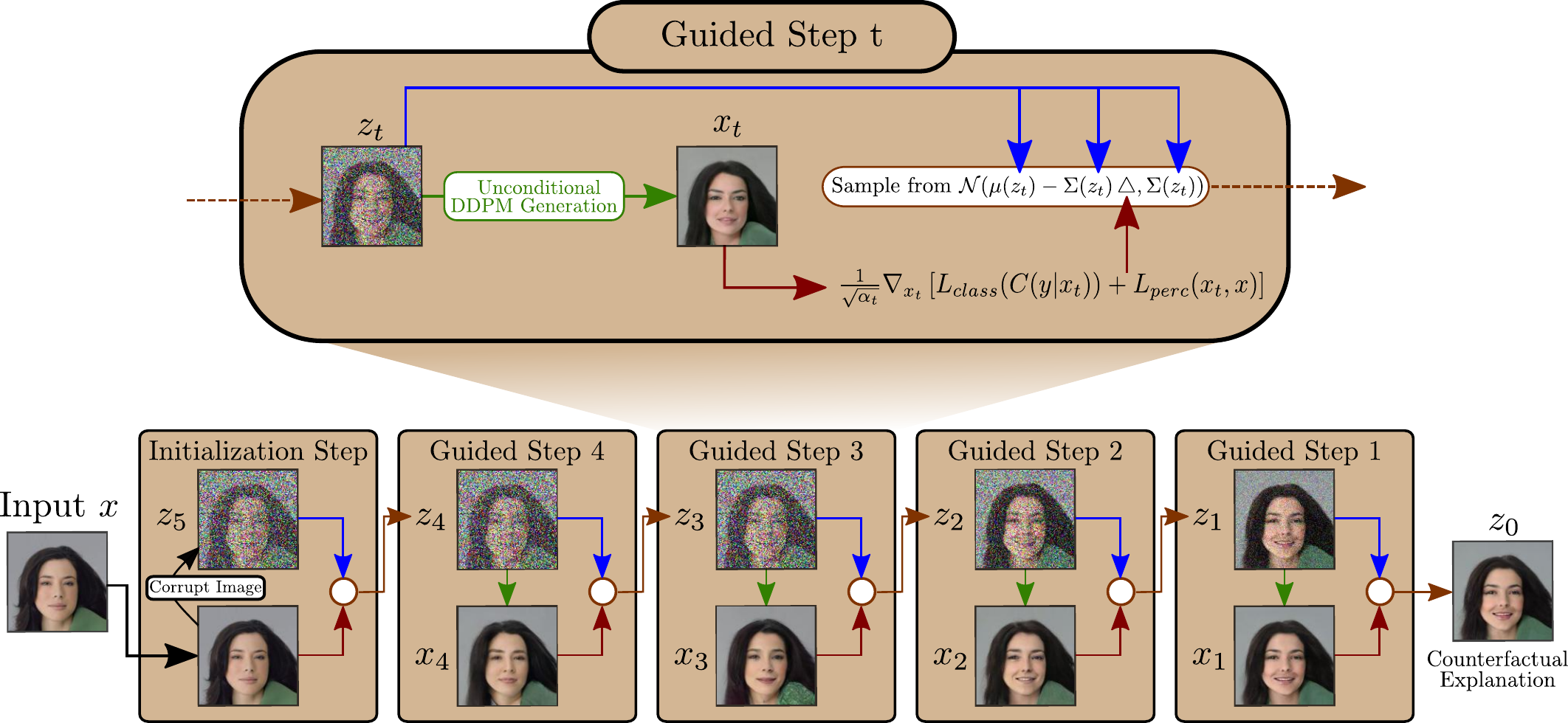}
    \caption{\textbf{DiME: \underline{Di}ffusion \underline{M}odels for Counterfactual \underline{E}xplanations.} Given an input instance $x$, we perturb it following Eq.~\ref{eq:degradation} to get $z_\tau$ (here $\tau=5$). 
    At time step $t$, we use the DDPM model to generate a clean image $x_t$ to obtain the clean gradient $L_{class}$ and $L_{perc}$ with respect to $x_t$. Finally, we sample $z_{t-1}$ using the guiding optimization process on Eq.~\ref{eq:generalized-diff}, using the previously extracted clean gradients.}
    \label{fig:baselinecf}
\end{figure*}

We take an image editing standpoint on CE generation, as illustrated Fig.~\ref{fig:baselinecf}. We start from a query image $x$. Initially, we rely on the forward process starting from $x_\tau=x$ to compute a noisy version $z_\tau$, with $1\leq\tau\leq T$. Then we go back in the reverse Markov chain using the guided diffusion (Eq~\ref{eq:generalized-diff}) to recover a counterfactual (hence altered) version of the query sample. Building upon previous approaches for CEs based on other generative models~\cite{Singla2020Explanation,wachter2018CounterfactualExplanationsOpening,jacob2021steex}, we rely on a loss function composed of two components to steer the diffusion process: a classification loss $L_{class}$, and a perceptual loss $L_{perc}$. The former guides the image edition into imposing the target label, and the latter drives the optimization in terms of proximity.

In the original implementation of the guided diffusion~\cite{Dhariwal2021DiffusionMB}, the loss function uses a classifier applied directly to the current noisy image $z_t$. This approach is appropriate since the considered classifier can make robust predictions under noisy observations, \textit{i.e.} it was trained on noisy images. Regardless, such an assumption on the classifier under scrutiny would imply a substantial limitation in the context of counterfactual examples.
We circumvent this obstacle by adapting the guided diffusion mechanism. To simplify the notations, let $x_t$ be the clean image produced by the iterative unconditional generation on Eq~\ref{eq:sampling} using as the initial condition $z_t$. In fact, this makes $x_t$ a \textit{function} of $z_t$ because we denoise $z_t$ recursively with the diffusion model $t$ times to obtain $x_t$. 
Luckily, we can safely apply the classifier to $x_t$ since it is not noisy. So, we express our loss as:
\begin{equation}\label{eq:loss-func}
    L(z_t;y, x) = \mathbb{E}[\underbrace{\lambda_{c} L_{class}(C(y|x_t)) + \lambda_{p} L_{perc}(x_t, x)}_{\tilde L(x_t; y, x)} ],
\end{equation}
where $C(y|x_t)$ is the posterior probability of the category $y$ given $x_t$, and $\lambda_c$ and $\lambda_p$ are constants.
Note that an expectation is present due to the stochastic nature of $x_t$. In practice, computing the loss gradient would require sampling several realizations of $x_t$ and taking an empirical average. We restrict ourselves to a single realization per step $t$ for computational reasons and argue that this is not an issue. Indeed, we can partly count on an averaging effect along the time steps to cope with the lack of individual empirical averaging. Besides, the stochastic nature of our implementation is, in fact, an advantage because it introduces more diversity in the produced CEs, a desirable feature as advocated by \cite{Rodriguez_2021_ICCV}.

Using this strategy, the dependence of the loss on $x_t$, rather than directly from $z_t$, renders the gradient computation more challenging.
Indeed, formally it would require to apply back-propagation from $x_t$ back to $z_t$: 
\begin{equation}\label{eq:clean-grads-general}
    \nabla_{z_t}L(z_t; y, x) = \left(\frac{Dx_t}{Dz_t}\right)^T\cdot\nabla_{x_t}\tilde L(x_t; y, x).
\end{equation}
Unfortunately, this computation requires retaining Jacobian information throughout the entire computation graph, which is very deep when $t$ is close to $\tau$. As a result, backpropagation is too memory intensive to be considered an option. To bypass this pitfall, we shall rely on the forward sampling process, which operates in a single stage (Eq.~\ref{eq:degradation}). Using the reparametrization trick~\cite{Kingma2014}, one obtains
\begin{equation}
    z_t = \sqrt{\alpha_t} x_t + \sqrt{1 - \alpha_t} \epsilon, \; \epsilon \sim \mathcal{N}(0, \mathbf{I}).
\end{equation}
Thus, by solving $x_t$ from $z_t$, we can leverage the gradients of the loss function with respect to the noisy input, a consequence of the chain rule. Henceforth, the gradients of $L$ with respect to the noisy image become
\begin{equation} \label{eq:clean-grads}
    \nabla_{z_t}L(z_t; y, x) = \frac{1}{\sqrt{\alpha_t}} \nabla_{x_t}\tilde L(x_t; y, x).
\end{equation}
This approximation is possible since the DDPM estimates the reverse Markov chain to fit the forward corruption process. Thereby, both processes are similar.

To sum up, Fig.~\ref{fig:baselinecf} depicts the generation of a counterfactual explanation with our algorithm: DiME. We start by corrupting the input instance $x=x_\tau$ following Eq.~\ref{eq:degradation} up to the noise level $t=\tau$. Then, we iterate the following two stages until $t=0$: \textit{(i)} First, using the gradients of the previous clean instance $x_{t-1}$, we guide the diffusion process to obtain $z_{t-1}$ using Eq.~\ref{eq:generalized-diff} with the gradients computed in Eq.~\ref{eq:clean-grads}. \textit{(ii)} Next, we estimate the clean image $x_t$ for the current time step $z_{t-1}$ with the unconditional generation pipeline of DDPMs. The final instance is the counterfactual explanation. If we do not find an explanation that fools the classifier under observation, we increase the constant $\lambda_c$ and repeat the process.

\textbf{Implementation Details.} 
To train the unconditional DDPM model, we used the publicly available code of~\cite{Dhariwal2021DiffusionMB}. We include all training and architectural details in the supplemental material. In practice, we incorporate additionally an $\ell_1$ loss, $\eta || z_t - x ||_1$, between the noisy image $z_t$ and the input $x$ to improve the $\ell_1$ metric on the pixel space. We empirically set $\eta$ small to avoid any significant impact on the quality of the explanations. Our diffusion model generates faces using 500 diffusion steps from the normal distribution. We re-spaced the sampling process to boost inference speed to generate images with 200 time-steps at test time. We use the following hyper-parameters settings:  $\lambda_{p}=30$, $\eta=0.05$, and $\tau=60$. Finally, we set $\lambda_{c} \in \{8, 10, 15\}$ to iteratively find the counterfactuals. We consider that our method failed if we do not find any explanation after exhausting the values of $\lambda_c$.
\section{Experiments}

\textbf{Experimental goals.} In this section, we evaluate DiME, our CE approach, using standard metrics. Also, we develop new tools to go beyond the current evaluation practices. Let us then recap the principles of current evaluation metrics, following previous works~\cite{Rodriguez_2021_ICCV,Singla2020Explanation}. The first goal of CEs is to create {\em realistic} explanations that {\em mislead} the classifier under observation. The capacity to change the classifier decision is typically exposed as a flip ratio (FR). Following the image synthesis research literature, the Frechet Inception Distance~\cite{NIPS2017_8a1d6947} (FID) measures the fidelity of the image distribution. The second goal of CE methods is to create proximal and sparse images. Among other tools, the XAI community adopted the Face Verification Accuracy~\cite{8373813} (FVA) and Mean Number of Attributes Changed (MNAC)~\cite{Rodriguez_2021_ICCV}. On the one hand, the MNAC metric looks at the face attributes that changed between the input image and its counterfactual explanation, disregarding if the individual's identity changed. On the other hand, the FVA looks at the individual's identity without considering the difference of attributes. 

Despite their usefulness, the previous metrics miss two important properties of CEs. Indeed, following \cite{Rodriguez_2021_ICCV}, to give a sense of trust in a classifier, the CEs must also produce diverse explanations and ensure that the classifier is not subject to spurious correlations. On the one hand, generating diverse explanations is useful to discover the brittleness of CNNs. Mothilal \textit{et al.}~\cite{Mothilal2020ExplainingML} propose a pair-wise distance to evaluate the diversity of counterfactual examples. Nevertheless, this work is exclusively dedicated to tabular data. We propose a simple adaptation to images based on the LPIPS metric \cite{Zhang_2018_CVPR_Unreasonable}. On the other hand, current assessments to detect spurious correlations (e.g., in \cite{Singla2020Explanation}) are quite extreme. They rely on modified datasets by entangling two attributes artificially to a full extent, e.g., all males are smiling, and all women are not. They also assume that in standard benchmarks, attributes are not entangled at all. Under this assumption, a classifier trained in this setting can be safely used as an oracle for the attributes, as proposed for computing MNAC. Actually, we show that this assumption can be largely erroneous and, therefore, challenge the derived metrics' validity. Based on our analysis, we designed a metric called {\em Correlation Difference} to assess if a counterfactual approach adequately reveals subtle ``spurious correlations'' (see Section \ref{sec:suriouscorel}).

\textbf{Dataset.} 
In this paper, we study the CelebA dataset~\cite{liu2015faceattributes}. Following standard practices, we preprocess all images to a $128\times128$ resolution. CelebA contains 200k images, labeled with 40 binary attributes. Previous works validate their methods on the \emph{smile} and \emph{young} binary attributes, ignoring all other features. Finally, the architecture to explain is a DenseNet121~\cite{Huang_2017_CVPR}. Given the binary nature of the task, the target label is always the opposite of the prediction. If the model correctly estimates an instance's label, we flip the model's forecast. Else, we modify the input image to classify the image correctly. 

\subsection{Realism, Proximity and Sparsity Evaluation} \label{sec:quant-val}

\begin{table}[t]
    \centering
    \begingroup
    \setlength{\tabcolsep}{3pt} % Default value: 6pt
    \renewcommand{\arraystretch}{1.1}
    \begin{tabular}{c|ccc|ccc} \toprule
                        & \multicolumn{3}{c|}{Smile}                      & \multicolumn{3}{c}{Young}                       \\ \midrule
        \textbf{Method} & \textbf{FID} ($\downarrow$) & \textbf{FVA} ($\uparrow$) & \textbf{MNAC} ($\downarrow$) & \textbf{FID} ($\downarrow$) & \textbf{FVA} ($\uparrow$) & \textbf{MNAC} ($\downarrow$) \\ \midrule
        xGEM+~\cite{Joshi2018xGEMsGE}          & 66.9         & 91.2         & -              & 59.5         & 97.5         & 6.70 \\
        PE~\cite{Singla2020Explanation}              & 35.8         & 85.3         & -              & 53.4         & 72.2         & 3.74 \\
        DiVE~\cite{Rodriguez_2021_ICCV}            & 29.4         & 97.3         & -                   & 33.8         & \textbf{98.2}& 4.58 \\
        DiVE$^{100}$            & 36.8        & 73.4         & 4.63                & 39.9          & 52.2         & 4.27 \\
        DiME              & \textbf{3.17}& \textbf{98.3}& \textbf{3.72}  & \textbf{4.15}& 95.3         & \textbf{3.13} \\\bottomrule
    \end{tabular}
    \endgroup
    \vspace{2mm}
    \caption{\textbf{State-of-the-Art results.} We compare our model performance against the State-of-the-Art on the FID, FVA and MNAC metrics. The values in \textbf{bold} are the best results. All metrics were extracted from~\cite{Rodriguez_2021_ICCV}. Our model has a 10 fold improvement on the FID metric. We extracted all results from Rodriguez \textit{et al.}' work~\cite{Rodriguez_2021_ICCV}.}
    \label{tab:main_results}
\end{table}

To compute the FID, the FVA, and the MNAC, we consider only those successful counterfactual examples, following previous studies~\cite{Rodriguez_2021_ICCV,Singla2020Explanation}. The FVA is the standard metric for face recognition. To measure this value, we used the cosine similarity between the input image and its produced counterfactual on the feature space of a ResNet50~\cite{He2016DeepRL} pretrained model on VGGFace2~\cite{8373813}. The instance and the explanation share the same identity if the similarity is higher than 0.5. So, the FVA is the mean number of faces sharing the same identity with their corresponding CE. To compute the MNAC, we fine-tuned the VGGFace2 model on the CelebA dataset. We refer to the fine-tuned model as the \emph{oracle}. Thus, the MNAC is the mean number of attributes for which the oracle switch decision under the action of the CE. For a fair comparison with the State-of-the-Art, we trained all classifiers, including the fine-tuned ResNet50 for the MNAC assessment, using the DiVE's~\cite{Rodriguez_2021_ICCV} available code.

DiVE do not report their flip rate (FR). This raises a concern over the fairness of comparing our methods. Since some metrics depend highly on the number of samples, especially FID, we recomputed their CEs. To our surprise, their flip ratio was relatively low (44.6$\%$ for the smile category). In contrast, we achieved a success rate of 97.6 and 98.9 for the smile and young attributes, respectively. Therefore, we calculated the counterfactual explanations with 100 optimization steps and reported the results as DiVE$^{100}$. DiVE$^{100}$'s success rates are $92.0\%$ for smile and $93.4\%$ for young, which is comparable to ours. 

We show DiME's performance in Table~\ref{tab:main_results}. Our method beats the previous literature in five out of six metrics. For instance, we have a $\sim$10 fold improvement on the FID metric for the smile category, while the young attribute has an $\sim$8 fold improvement. We credit these gains to the generative capabilities of the diffusion model. Further, our generation process does not require entirely corrupting the input instance; hence, the coarse details of the image remain. The other methods rely on latent space-based architectures. Thus, they require to compact essential information removing outlier data. Consequently, the generated CEs cannot reconstruct the missing information, losing significant visual components of the image statistics. 

Despite the previous advantages, we cannot fail to notice that DiME is less effective in targetting the young attribute than the smile.
The smile and young attributes have distinct features. 
The former is delineated by localized regions, while the latter scatters throughout the entire face.
Thus, the gradients produced by the classifier differ between the attributes of choice; for the smile attribute, the gradients are centralized while they are outspread for the young attribute. 
We believe that this subtle difference underpins the slight drop of performance (especially with respect to FVA) in the young attribute case. 
This hypothetical explanation should be confirmed by a more systematic study of various attributes, though this phenomenon is out of scope of the paper.

\subsection{Diversity Assessment}

One of the most crucial traits of counterfactual explanations methodologies is the ability to create multiple and diverse examples~\cite{Rodriguez_2021_ICCV,Mothilal2020ExplainingML}. As stated in the methodology section, DiME's stochastic properties enable the sampling of diverse counterfactuals. To measure the capabilities of different algorithms to produce multiple explanations, we computed the mean pair-wise LPIPS~\cite{Zhang_2018_CVPR_Unreasonable} metric between five independent runs. A higher LPIPS means increased perceptual dissimilarities between the explanations, hence, more diversity. To compute the evaluation metric, we use all counterfactual examples, even the unsuccessful instances, because we search the capacity of exploring different traits. Note that we exclude the input instance to compute the metric since we search for the dissimilarities between the counterfactuals. We compared DiME's performance with DiVE$^{100}$ and its Fisher Spectral variant on a small partition of the validation subset.

\begin{figure}[t]
    \centering
    \includegraphics[width=0.83\textwidth]{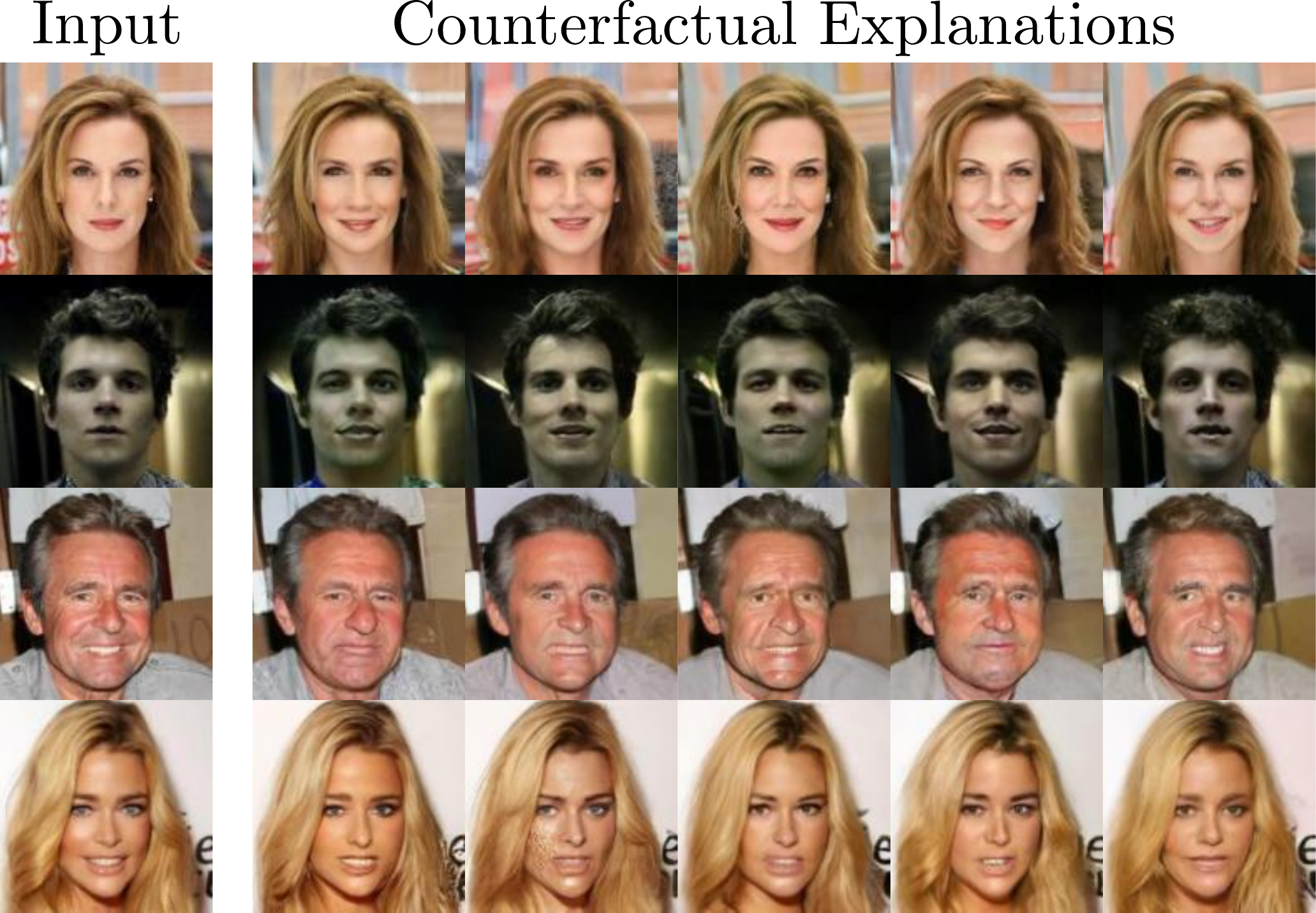}
    \caption{\textbf{Diversity Counterfactual examples.} The classifier predicts first two input images as non-smiley and the last two as smiley. In this example all explanations fool the classifier. Our CE pipeline is capable of synthesising diverse counterfactuals without any additional mechanism.}
    \label{fig:diversity}
\end{figure}

We visualize some examples in Fig.~\ref{fig:diversity} and show the performance of the five runs on the supplementary material. We obtained an LPIPS value of 0.213. In contrast, DiVE~\cite{Rodriguez_2021_ICCV} and its Spectral Fisher variant obtained an LPIPS of 0.044 and 0.086, respectively. Recall that DiME does not have an explicit mechanism to create diverse counterfactuals. Its only mechanism is the stochasticity within the sampling process (Eqs.~\ref{eq:sampling} and~\ref{eq:generalized-diff}). In contrast, DiVE relies on a diversity loss when optimizing the eight explanations. Yet, our methodology achieves higher diversity with the LPIPS metric even without an explicit mechanism.

\subsection{Discovering Spurious Correlations \label{sec:suriouscorel}}

\begin{figure}[t]
    \centering
    \includegraphics[width=\textwidth]{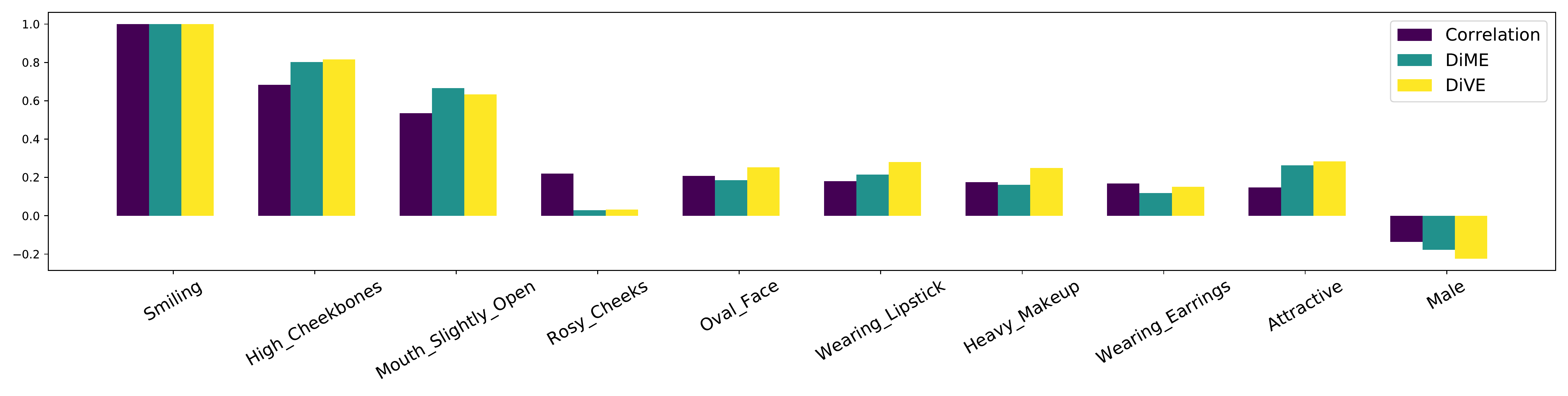}
    \caption{\textbf{Spurious Correlation Detection.} We show the top 9 most correlated attributes in the label space with ``smile''. We obtained the Pearson Correlation Coefficient from the ground truth on the training set. Albeit the difference in the MNAC performance, DiME and DiVE achieve to detect the spurious correlations similarly. We show all the remaining attributes in the supplementary material.}
    \label{fig:correlation}
\end{figure}

The end goal of counterfactual examples is to uncover the modes of error of a target model. Current evaluation protocols~\cite{Singla2020Explanation} search to assess the spurious correlations by inducing artificial entanglements between two supposedly uncorrelated attributes. Conventionally, the experiment involves mixing the smile and gender attributes. The goal is then to evaluate whether or not the CE algorithm is able to reveal the correlation. 
To assess this capability, it is common to verify if both the target and the entangled attributes change when producing the CEs. In our opinion, such an extreme experiment does not shed light on the ability to reveal spurious correlations in real situations. Indeed, the considered configuration assumes that only two labels are entangled and that this entanglement is complete.

In fact, as depicted in Fig.~\ref{fig:correlation}, in real datasets such as CelebA, many labels are correlated at multiple levels. As a result, this phenomenon calls the previously proposed correlation experiment into question. It also raises concerns about the value of the MNAC metric, or measurement tools such as the LVS~\cite{Hvilshoj2021OnQE}. As a matter of fact, while small MNAC values are often considered desirable (see \cite{Rodriguez_2021_ICCV,Singla2020Explanation,wachter2018CounterfactualExplanationsOpening}), the presence of spurious correlations challenges this interpretation. Indeed, consider the following illustrating scenario comparing two CE algorithms: the first one exposes all spurious correlations correctly; the second one can solely edit the main feature. Since the first method produces many changes, it will display a \textit{high} MNAC while the alternate algorithm reaches a \textit{low} MNAC value. This false sense of high performance does not reflect the true accomplishment of the first model: detecting spurious correlations. So, we propose to amend the MNAC measurement into a new metric called the Correlation Difference (CD), more adapted to assess the capacity of CEs to reveal spurious correlations.

The goal of CD is to measure the difference between the true correlations and the changes produced by the explanations. To achieve this, let $c_{qa}$ be the true correlation computed on the label space between the attribute labels $q$ and $a$. To measure the correlations in the prediction space, first we define
\begin{equation}
    \delta_a(x) = O_a(M_q(x)) - O_a(x),
\end{equation}
where $O_a(\cdot)$ is the oracle's binary prediction of its input for the attribute $a$, $M_q$ is the counterfactual method targeting the query attribute $q$ on an image $x$. This measure looks at the signed changes implied by $M_q$ on $x$. So, now we can measure the relative changes on the attributes when computing a counterfactual example. Therefore, we can calculate the correlation coefficient $c_{qa}^{M}$ between $\{\delta_q(x)\}_x$ and $\{\delta_a(x)\}_x$ to compare it with $c_{qa}$, the true correlation\footnote{The series $\{\delta_q(x)\}_x$ and $\{\delta_a(x)\}_x$ uses solely the successful counterfactual explanations. Further, we used the Pearson Correlation Coefficient to compute $c_{qa}$ and $c_{qa}^{M}$.}. Accordingly, CD is:
\begin{equation}
    CD = \sum_{a} | c_{qa} - c_{qa}^{M} |.
\end{equation}

We apply our proposed metric on DiME and DiVE$^{100}$'s counterfactual explanations. We got a CD of 2.30 while DiVE$^{100}$ 2.36 on CelebA's validation set, meaning that DiVE$^{100}$ lags behind DiME. However, the margin between the two approaches is only slender. This reveals our suspicions: the MNAC results presented in Table~\ref{tab:main_results} give a misleading impression of a robust superiority of DiME over DiVE$^{100}$.

\subsection{Impact of the noise-free input of the classifier}\label{sec:ablations}

In this section, we assess the impact of our main adjustment over the original guided diffusion process. Recall that we argued that it is important to apply the classifier on noise-free images $x_t$ and not on the current noisy version $z_t$ in order to obtain a robust gradient direction. To validate this claim, we compare our approach to an alternative, dubbed \emph{Direct}. It uses the gradient of the classifier applied directly to the noisy instance $z_t$. In this case, we removed the constant $\nicefrac{1}{\sqrt{\alpha_t}}$ since we compute directly the gradients with respect to $z_t$. To complete the picture, we also consider two additional variations of our approach. The first one, called \emph{Naive}, uses the gradient of the input image at each time step to guide the optimization process. Therefore, it is not subject to noise issues, but it disregards the guidance that was already applied until time step $t$. The second variation is a near duplicate of DiME except for the fact that it ends the guided diffusion process as soon as $x_t$ fools the classifier. We name this approach \emph{Early Stopping}. Eventually, we will also evaluate the DDPM generation without any guiding and beginning from the corrupted image at time-step $\tau$  to mark a reference of the performance of the DDPM model.   

% \begin{table}[t]
%     \centering
%     \begingroup
%     \setlength{\tabcolsep}{4pt} % Default value: 6pt
%     \renewcommand{\arraystretch}{1.1}
%     \begin{tabular}{c|c c c c c} \toprule
%         \textbf{Method} & \textbf{FID}   & \textbf{FID}$^*$ & \textbf{FR}  & \textbf{$\ell_1$} & \textbf{BKL}\\ \midrule
%         Direct          & 50.51          & 50.51            & 19.7         & 0.0454            & 0.297 \\ 
%         Naive           & 75.42          & 98.93 $\pm$ 2.36 & 70.0         & 0.0624            & 0.139 \\
%         Unconditional   & 24.03$^\S$     & 53.22$^\S$ $\pm$ 0.98& 8.6          & 0.0492            & 0.265\\ \midrule
%         Ours            & \textbf{20.51} & 50.20 $\pm$ 1.00 & \textbf{97.9}& \textbf{0.0430}   & \textbf{0.076}\\\bottomrule
%     \end{tabular}
%     \endgroup
%     \vspace{2mm}
%     \caption{\textbf{Standard method \textit{vs} ``Naive'' methods.} In this table we show the result of incorporating the clean gradients generated with the unconditional generation to guide each step on the reverse diffusion algorithm. We refer as naive using only the gradients of the input image. The direct method is using the gradients of the noisy instance directly. Due to the low number of instances for the FID$^*$ metric, we computed it using 10 times and show their mean and standard deviation. $^\S$ FID computed with all instances, even unsuccessful counterfactuals.}
%     \label{tab:naive}
% \end{table}

\begin{table}[t]
    \centering
    \begingroup
    \setlength{\tabcolsep}{4pt} % Default value: 6pt
    \renewcommand{\arraystretch}{1.1}
    \begin{tabular}{c|c c c c} \toprule
        \textbf{Method} & \textbf{FR} ($\uparrow$)         & \textbf{FID}$^+$($\downarrow$) & \textbf{$\ell_1$}($\downarrow$)  & \textbf{BKL}($\downarrow$)\\ \midrule
        Direct          & 19.7                             & \textbf{50.51}                 & 0.0454                           & 0.297 \\
        Naive           & 70.0                             & 98.93 $\pm$ 2.36               & 0.0624                           & 0.139 \\
        Early Stopping  & 97.3                             & 51.97 $\pm$ 0.77               & 0.0467                           & 0.350 \\
        Unconditional   & 8.6                              & 53.22$^\S$ $\pm$ 0.98          & 0.0492                           & 0.265\\ \midrule
        DiME            & \textbf{97.9}                    & \textbf{50.20} $\pm$ 1.00      & \textbf{0.0430}                  & \textbf{0.076}\\\bottomrule
    \end{tabular}
    \endgroup
    \vspace{2mm}
    \caption{\textbf{DiME \textit{vs} ``Naive'' variations.} This table shows the advantages of the proposed adjustment to incorporate the classifier under observation. We clearly see that including the clean gradients benefits DiME on all metrics, especially the FR.
    %%In this table we show the result of incorporating the clean gradients generated with the unconditional generation to guide each step on the reverse diffusion algorithm. We refer as naive using only the gradients of the input image. The direct method is using the gradients of the noisy instance directly. Due to the low number of instances for the FID$^+$ metric, we computed it using 10 times and show their mean and standard deviation. $^\S$FID$^+$ computed with all instances, even unsuccessful counterfactuals.
    }
    \label{tab:naive}
\end{table}

\emph{Notes on considered metrics:}
In addition to FID, FR and $\ell_1$ metrics, we also evaluate the following metric: $BKL(y||M_q(x)) = 1 - C(y | M_q(x))$. It is the complement of the target label's probability, but whose origin is a bounded remapping of a KL divergence, hence the notation $BKL$. A low BKL means that the classifier under observation classifies the counterfactual example $M_q(x)$ with high confidence and is effectively fooled by the CE. 

Also, given that many variants are considered, we created a small and randomly selected mini-val to evaluate the various metrics. Besides, given that the different baselines can display varying levels of FR, we condition the FID computation on the successful CEs only. However, it is well known that FID is strongly biased, especially when using a low number of samples. To mitigate this bias, we use the same number of CEs for each baseline (the least number of successful CEs) and repeat this computation $10$ times to report a mean and standard deviation. We denote this fair FID as FID$^+$. Similarly, we compute the $\ell_1$ and BKL solely for successful counterfactuals. 

We show the results of the different variations in Table~\ref{tab:naive}. The most striking point is that when compared to the Naive and Direct approaches, the unimpaired version of DiME is the most effective in terms of FR by a large margin. This observation validates the need for our adjustment of the guided diffusion process. Further, our approach is also superior to all other variations in terms of the other metrics. At first glance, we expected the unconditional generation to have better FID than DiME and the ablated methods. However, we believe that the perceptual part of our loss is beneficial in terms of FID. Therefore, the unconditional FID is higher. Similarly, the early stopping variant is also impacted in terms of FID and BKL because the optimization is brought to an end prematurely.

We complement this ablation study in the supplementary material. In particular, we explore the role of the initial noise level $\tau$ and the mixing factor $\lambda_c$. Other quantitative and qualitative results are presented therein.
%To complement this ablation study, we explore the variables that affect the generation of counterfactual explanations in the supplementary material. In addition, we included the results of varying the initial noise level $\tau$ and the gradient's scale $\lambda_c$. Finally, we analyzed the series of clean images produced each time step. 

%% \input{sections/ablations/b_scale}
%% \input{sections/ablations/c_init}
%\input{sections/ablations/d_distribution}

\subsection{Qualitative Results}

\begin{figure}[t]
    \centering
    \includegraphics[width=0.93\textwidth]{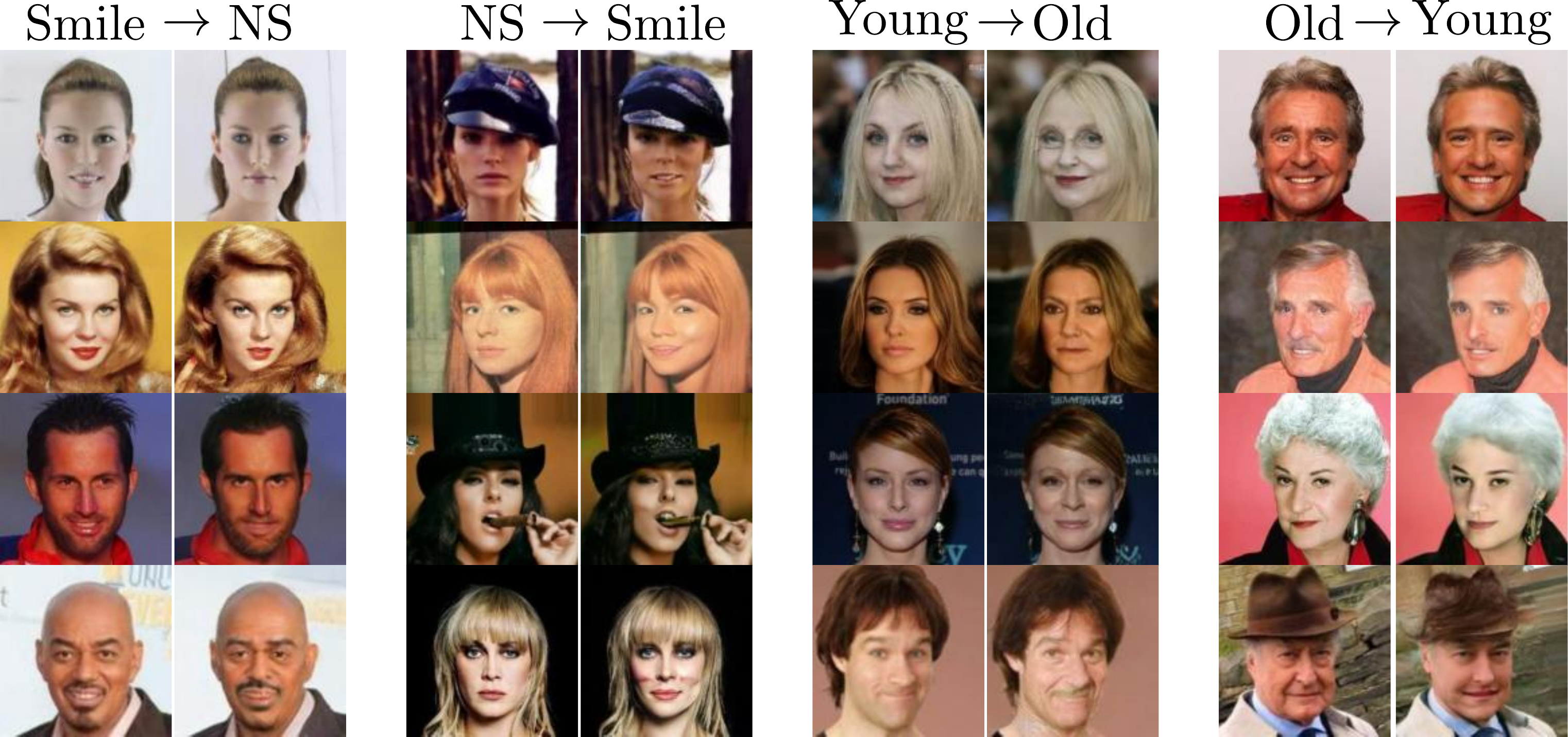}
    \caption{\textbf{Qualitative Results.} We visualize some images and its corresponding counterfactual explanation produced by our proposed approach. Our methodology achieves to incorporate small but perceptually tangible changes in the image. NS stands for Non-Smiley.}
    \label{fig:qualitative}
\end{figure}

We visualize some inputs (left) and the counterfactual examples (right) produced by DiME in Fig.~\ref{fig:qualitative}. We show visualizations for the attributes smile and young, yet we will include visualizations for other categories in the supplementary material. At first glance, the results reveal that the model performs semantical editings into the input image. In addition, uncorrelated features and coarse structure remain almost unaltered. We observe slight variations on some items, such as the pendants, or out-of-distribution shapes such as hands. DiME fails to reconstruct the exact shape of these objects, but the essential aspect remains the same.

\subsection{Limitations}

Our pipeline for counterfactual explanation has several limitations. Although we show the benefits of using our model to generate CEs, we are far from accomplishing all aspects crucial for the XAI community. First of all, our method is slow and computationally demanding. Since we are using DDPMs, we adopt most of their limitations. For instance, we need to use the DDPM model $\sim$1800 times to generate a single explanation. This aspect is undesired whenever the user requires an explanation on the fly. Finally, we require access to the training data. This limitation is common in many previous studies. However, this aspect is vital in fields where data is sensitive. Although access to the training data is permitted in many cases, we restrict ourselves to using the data without any labels. 
\section{Conclusion}

In this paper, we explore the novel diffusion models in the context of counterfactual explanations. By harnessing the conditional generation of the guided diffusion, we achieve successful counterfactual explanations through DiME. These explanations follow the requirements given by the XAI community: they produce a small but tangible change in the image while remaining realistic.
The performance of DiME is confirmed based on a battery of standard metrics.
DiME also exhibits strong diversity in the produced explanation. This is partly inherited from the intrinsic features of diffusion models, but it also results from a careful design of our approach.
Further, we show that the current approach to validate the sparsity of CE has significant conflicts with the assessment of spurious correlation detection. Finally, our proposed metric, Correlation Difference, correctly measures the impact of measuring the subtle correlation between labels. We hope that our work opens new ways to compute and evaluate counterfactual explanations. 

\subsection*{Acknowledgements} 
Research reported in this publication was supported by the Agence Nationale pour la Recherche (ANR) under award number ANR-19-CHIA-0017.

\bibliographystyle{splncs04}
\bibliography{egbib}

\clearpage

\section*{Supplementary material}

\appendix

\section{Implementation Details}

\textbf{DDPM Architectural and Training Details.}
We trained the unconditional DDPM using the publicly available code of~\cite{Dhariwal2021DiffusionMB}. 
Our model has the same architecture as the ImageNet's Unconditional DDPM of~\cite{Dhariwal2021DiffusionMB}, except for two differences. 
\textit{(i)} The number of diffusion steps for~\cite{Dhariwal2021DiffusionMB} is $1000$ while we use $500$ steps only. 
\textit{(ii)} we reduced the number of inner channels from 256 to 128 given that CelebA's complexity is far lower than ImageNet's. 
We trained our model for 270.000 iterations with a batch size of 75 on 5 GPUs, \textit{i.e.} a batch size of 15 per GPU.
We set the learning rate to $1\times 10^{-4}$ with a weight decay of 0.05 and no dropout. 
Although we selected this configuration for the architecture and the training, we did not perform an exhaustive exploration since we are not searching to evaluate the diffusion model performance. 

\textbf{Loss selection.}
The selection of the losses influences the convergence of the stochastic optimization process for the CE. 
We chose the standard VGG19 perceptual loss as the $L_{perc}$ loss. 
For the classification loss $L_{class}$, we opted to maximize directly logits of the target class instead of the log probability. 
More specifically, we minimize the negative logits. 

\begin{table}[b]
    \centering
    \begingroup
    \setlength{\tabcolsep}{4pt} % Default value: 6pt
    \renewcommand{\arraystretch}{1.1}
    \begin{tabular}{c|c c c c} \toprule
        \textbf{Seed} & \textbf{FID}($\downarrow$)  & \textbf{FR}($\uparrow$) & \textbf{$\ell_1$}($\downarrow$) & \textbf{BKL}($\downarrow$)\\ \midrule
        1             & 20.51         & 97.9        & 0.0430            & 0.076 \\
        2             & 20.60         & 97.6        & 0.0430            & 0.073 \\ 
        3             & 20.72         & 97.9        & 0.0431            & 0.067 \\
        4             & 20.67         & 97.7        & 0.0431            & 0.073 \\
        5             & 20.46         & 98.2        & 0.0430            & 0.076 \\\bottomrule
    \end{tabular}
    \endgroup
    \vspace{2mm}
    \caption{\textbf{Diversity experiments.} We ran our method five times, varying the initial seed. The results show that our method is robust to the initial conditions, although the visual elements vary significantly.}
    \label{tab:diversity}
    \vspace{-0.4cm}
\end{table}

\section{Variability Evaluation}

We report the performances of the five different runs in Table~\ref{tab:diversity}. 
Even when we set different initial conditions for each iteration, DiME is robust to many instantiations. 
We visualize more images for the variability in section~\ref{seq-a:qualitative}. 
Many results vary significantly, yet DiME solves the counterfactuals in most cases.

\section{Correlation Discovery}

\begin{figure}[t]
    \centering
    \includegraphics[width=1\textwidth]{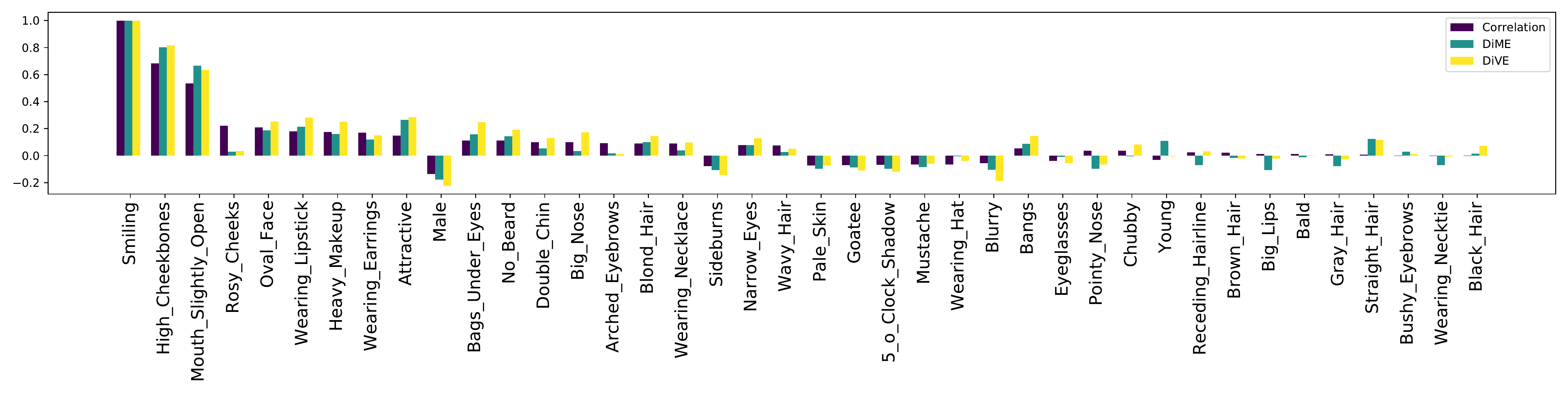}
    \caption{\textbf{Correlation discovery.} We visualize all the correlation discovered by DiME and DiVE for all attributes on CelebA.}
    \label{fig:cd-all}
\end{figure}

In section~4.3 of the main manuscript we discussed the importance of our proposed metric CD. 
Nevertheless, we visualize only the top 9 attributes given that the other attributes are far less correlated. 
For the sake of completing the study, we added the rest of attributes on Fig.~\ref{fig:cd-all}. Similarly, we observe that DiME and DiVE have similar capabilities finding correlations in the data.

\section{Ablations Studies}

To complement the ablation analysis on the components of DiME, we explore the variables that affect the generation of counterfactual explanations. 
On the one side, we analyze the impact of calibrating the gradients' scale. 
On the other side, we study the effects of varying the initial noise level, considering that adding more noise implies removing more details. 
Finally, we visualize the evolution of the clean images produced at each time step of the guiding process. %We visualize some results for all ablations of section~\ref{seq-a:qualitative}.

\subsection{Gradients' Scale Ablation}

\begin{table}[t]
    \centering
    \begingroup
    \setlength{\tabcolsep}{4pt} % 4pt
    \renewcommand{\arraystretch}{1.1} % 1.2
    \begin{tabular}{c|c c c c} \toprule
        $\lambda_c$       & \textbf{FID}$^+$($\downarrow$)  & \textbf{FR}($\uparrow$)  & $\mathbf{\ell_1}$($\downarrow$) & \textbf{BKL}($\downarrow$)\\ \midrule
        8                             & 22.93             & 80.1         & \textbf{0.0427}   & 0.058 \\
        10                            & 23.32             & 88.0         & 0.0432            & 0.041 \\ 
        15                            & 25.87             & 97.7         & 0.0446            & \textbf{0.019} \\ \midrule
        DiME                          & \textbf{22.48}    &\textbf{97.9} & 0.0430            & 0.076 \\\bottomrule
    \end{tabular}
    \endgroup
    \vspace{2mm}
    \caption{\textbf{Gradient Scales.} We show the impact of different scales choices. We computed the BKL and FID metrics solely from the successfully counterfactual explanations. Increasing the gradient scale $\lambda_c$ decreases the FID and $\ell_1$. From the Flip Ratio results, we see that most explanations are produced with a low scale value, hence producing similar results in the pixel space with high fidelity. Harder instances require the use of an increased scale to successfully produce the counterfactual example. The FID$^+$ is computed taking the the same number of samples for the experiment $\lambda_c=8$.}
    \label{tab:gradients}
    \vspace{-0.4cm}
\end{table}

The work Dhariwal and Nichol~\cite{Dhariwal2021DiffusionMB} ablates the scale of the classifier. 
Their results indicate a positive correlation between the quality of the images, measured with the FID, and the gradients' scale. 
So, we perform a similar study; to find a counterfactual explanation, we optimize the image formation with three different scales $\lambda_{c} \in \{8, 10, 15\}$ and choose the generated image with the smallest scale. 
Therefore, we analyze the individual contribution of each scale. 

We report the results in Table~\ref{tab:gradients}. 
In opposition to \cite{Dhariwal2021DiffusionMB}, we observe that when increasing the scale, the quality as measured by FID drops; more precisely the FID value increases.
This inconsistency with the observation of \cite{Dhariwal2021DiffusionMB} remains to elucidate. 
Although it is out of the scope of our work, we can at least point out a few potential sources of discrepancy.
First, the type of image edition that we perform is fundamentally different from the one considered in \cite{Dhariwal2021DiffusionMB}. 
In their work, the task correspond to generate an image conditionally to a categorical label. This categorical, hence discrete, aspect of the condition may be at odd with soft constraints (i.e. small gradient scale). It is not present in our context. 
Among other differences, one can note that we start the denoising process from an intermediate step $\tau\leq T$ while they start from the very last step $T$.
Eventually, we have a specific way of computing the gradient.

% We report the results in Table~\ref{tab:gradients}. 
% In opposition to~\cite{Dhariwal2021DiffusionMB}, we observe that when increasing the scale, the quality as measured by FID drops; more precisely, the FID value increases. 
% Nevertheless, we cannot conclude that this relationship results from instantiating the noisy image at an intermediate time step, thus avoiding corrupting essential coarse details. 
% This phenomenon is mostly due to both~\cite{Dhariwal2021DiffusionMB}, and our experimental setups vary significantly; for instance, there are different image domains (natural \textit{vs} face images), gradient extraction algorithms, and ground-truth labels. 
% we can comment this last paragraph if we liked more the last one.

In our particular context for CEs, we noticed a trade-off between the success rate and the quality of the images. 
Since we seek to produce sparse modifications, we created most explanations using the lowest scale. 
Thus, we enjoy the benefits of high-quality images. 
Further, we boost the FR by using higher scales at the cost of lowering the quality. 
Adding too much gradient produces out-of-distribution noise. 
The DDPM cannot recognize this noise, and therefore it produces artifacts on the image. 
However, these artifacts may coincide with patterns that impact the classifier response. 
%Finally, the difference in FID is due to different instantiations of the counterfactuals. 

\subsection{Initial Step Ablation}

% \begin{table}[t]
%     \centering
%     \begingroup
%     \setlength{\tabcolsep}{4pt} % Default value: 6pt
%     \renewcommand{\arraystretch}{1.1}
%     \begin{tabular}{c|c c c c c} \toprule
%         \textbf{Steps} & \textbf{FID} & \textbf{FID}$^*$ &\textbf{FR}   & $\mathbf{\ell_1}$& \textbf{BKL}\\ \midrule
%         50            & \textbf{20.19}& \textbf{20.19}   & 92.4         & \textbf{0.0406}   & 0.100 \\
%         60            & 20.51         & 20.94            & 97.9         & 0.0430            & 0.076 \\ 
%         70            & 22.39         & 23.21            & \textbf{99.7}& 0.0479            & \textbf{0.048} \\\bottomrule
%     \end{tabular}
%     \endgroup
%     \vspace{2mm}
%     \caption{\textbf{Initialization Step.} We show the result of different $\tau$ choices. The BKL and FID metrics are computed solely from the successfully counterfactual explanations. Using $\tau=60$ provides the best trade-off between image quality, Flip Ratio and similarity. The FID$^*$ is computed taking the the same number of samples for the experiment with less instances.}
%     \label{tab:steps}
% \end{table}

\begin{table}[t]
    \centering
    \begingroup
    \setlength{\tabcolsep}{4pt} % Default value: 6pt
    \renewcommand{\arraystretch}{1.1}
    \begin{tabular}{c|c c c c} \toprule
        \textbf{Steps} & \textbf{FID}$^+$($\downarrow$) &\textbf{FR}($\uparrow$)   & $\mathbf{\ell_1}$($\downarrow$)& \textbf{BKL}($\downarrow$)\\ \midrule
        50             & \textbf{20.19}   & 92.4         & \textbf{0.0406}   & 0.100 \\
        60             & 20.94            & 97.9         & 0.0430            & 0.076 \\ 
        70             & 23.21            & \textbf{99.7}& 0.0479            & \textbf{0.048} \\\bottomrule
    \end{tabular}
    \endgroup
    \vspace{2mm}
    \caption{\textbf{Initialization Step.} We show the result of different $\tau$ choices. The BKL and FID metrics are computed solely from the successfully counterfactual explanations. Using $\tau=60$ provides the best trade-off between image quality, Flip Ratio and similarity. We computed the FID$^+$ taking the the same number of samples for the experiment with less instances ($\tau=50$).}
    \label{tab:steps}
    \vspace{-0.4cm}
\end{table}
Following the previous experiment, we seek to find the main variables to produce valid yet sparse CEs. 
The other variable of interest is the initialization step $\tau$. 
On the one hand, a higher $\tau$ opens more opportunities to modify the image. But on the other hand, this increased generation power can be detrimental to 
the resulting image quality.
We report the results in Table~\ref{tab:steps}. 
The hyperparameter $\tau$ has a similar effect to the gradient scale $\lambda_c$; we observe a negative correlation between $\tau$ and the FID, and a positive one between $\tau$ and the FR. 
The image generation has more optimization steps when increasing the initial noise level. 
Thus, it easily reaches a counterfactual that fools the classifier at the cost of decreasing the CE sparsity, an unwanted effect in the CE community. 
Similarly, a low $\tau$ increases the sparsity, but the CEs are not as successful. 
Choosing $\tau=60$ finds an optimal equilibrium between both factors.

\subsection{Distribution Overtime}

Our pipeline uses the unconditional DDPM to enable the use of the classifier under observation. 
At each step, the classifier uses the generated image to compute the gradient with respect to the target label. 
Therefore, this image gives information on the optimization process at each time step. 
Hence, in this experiment, we explore the behavior of these images at each stage of the denoising process. 
We plot the probability of the target class given by the inspected architecture at each step to accomplish this. 

\begin{figure}[t]
    \centering
    \includegraphics[width=0.90\textwidth]{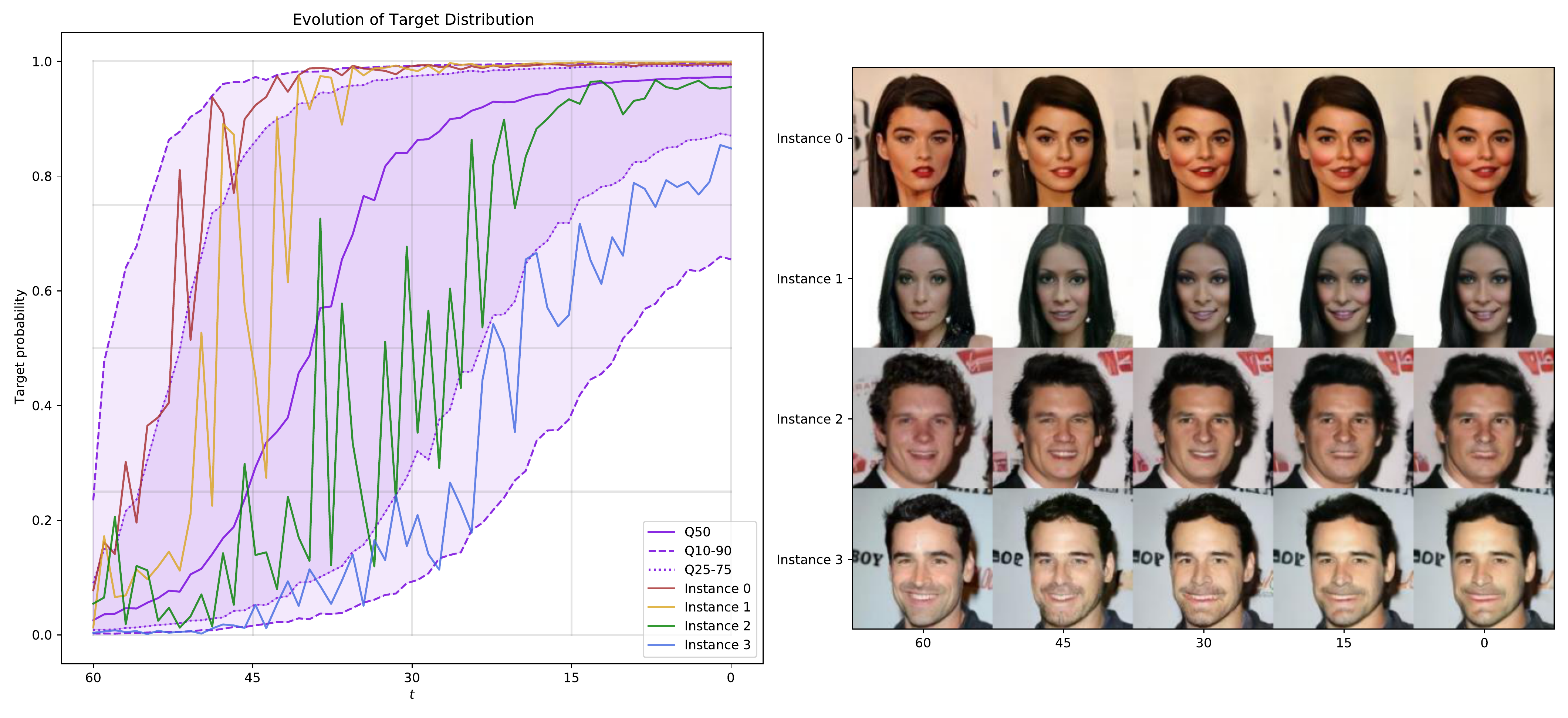}
    \caption{\textbf{Target distribution at each time-step.} We visualize the evolution of the target labels' probability. Each purple line represents a quantile of the probabilities. The color curves are cases shown on the right. In expectancy, the clean image at each time step increases. Nonetheless, typically the curves are sporadic. Yet, we observe an increase.}
    \label{fig:dist_on_time}
    \vspace{-0.4cm}
\end{figure}

In Fig.~\ref{fig:dist_on_time} we visualize the evolution of the target labels' probability over time, along with some examples. 
We see that the probability increases overtime on average. 
Nevertheless, the example instances show sporadic and non-steady development. 
Yet, we still observe an ascending behavior. 
Near the first steps, we see the most unstable conduct. 
Nonetheless, the optimization begins to settle when reaching a time steps near $0$ (approx. at $t=20$). 
We attribute this behavior to an averaging effect along time; when the image generation reaches the final steps, the variance nearly vanishes. 
Hence, the unconditional generation does not vary much, reaching an equilibrium. 
This observation relates to the comments of Equation~5 in the main text, where we argue for using a single realization of $x_t$ at each time step.
As mentioned in the main text, the absence of averaging at every step is partly mitigated in terms of the optimization objective by an averaging effect over time. 
But thanks to the randomness inherited from the early steps ($t\approx \tau$), the overall CE creation process still displays some diversity.

\section{Qualitative Results} \label{seq-a:qualitative}

\begin{figure}
    \centering
    \includegraphics[width=0.9\textwidth]{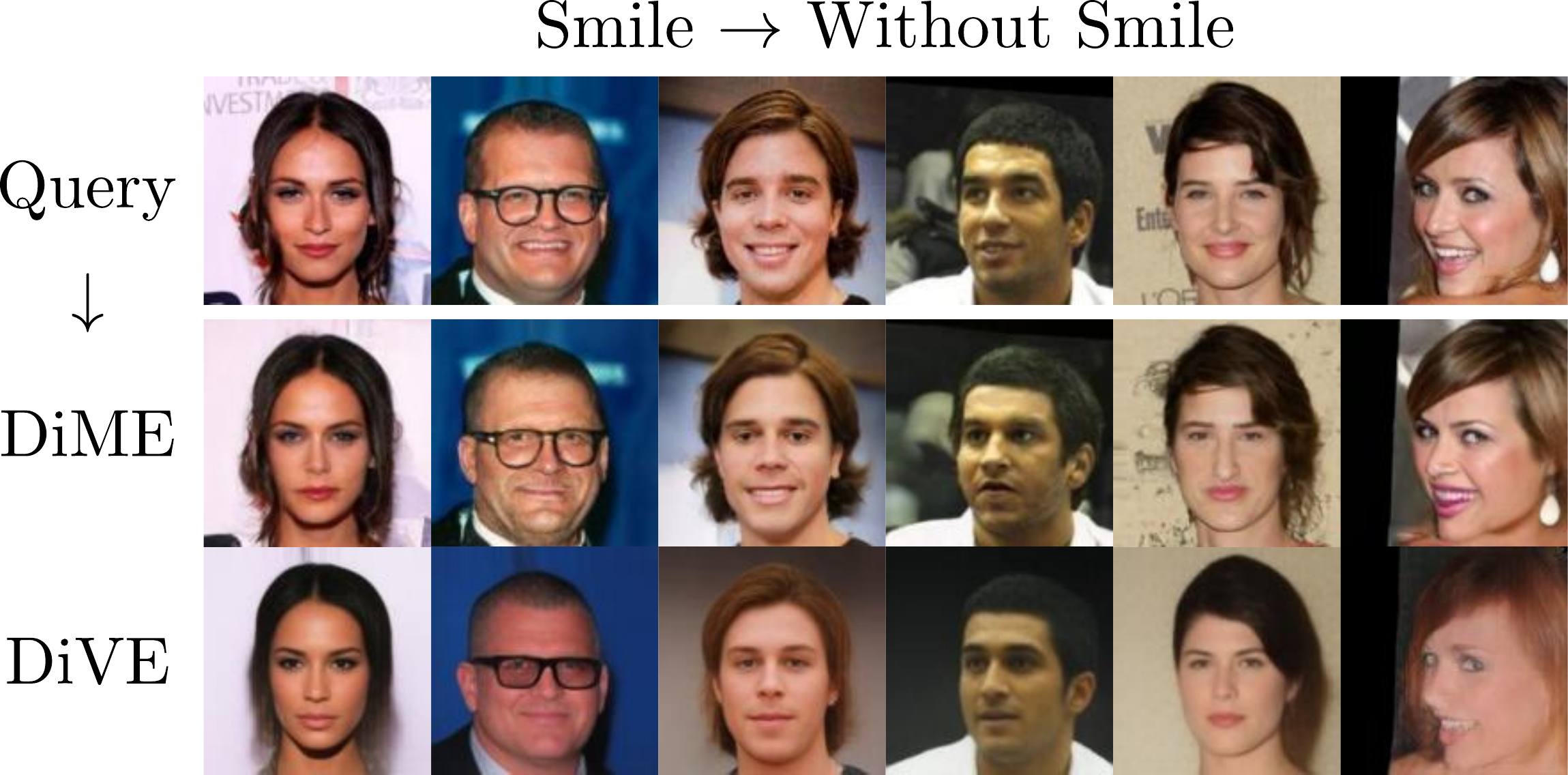}
    \caption{\textbf{Counterfactual explanations for the \textit{Smile}.} We visualize DiME and DiVE explanations targeting the label Smile.}
    \label{fig:s-t-ns}
    \vspace{-0.4cm}
\end{figure}

\begin{figure}
    \centering
    \includegraphics[width=0.9\textwidth]{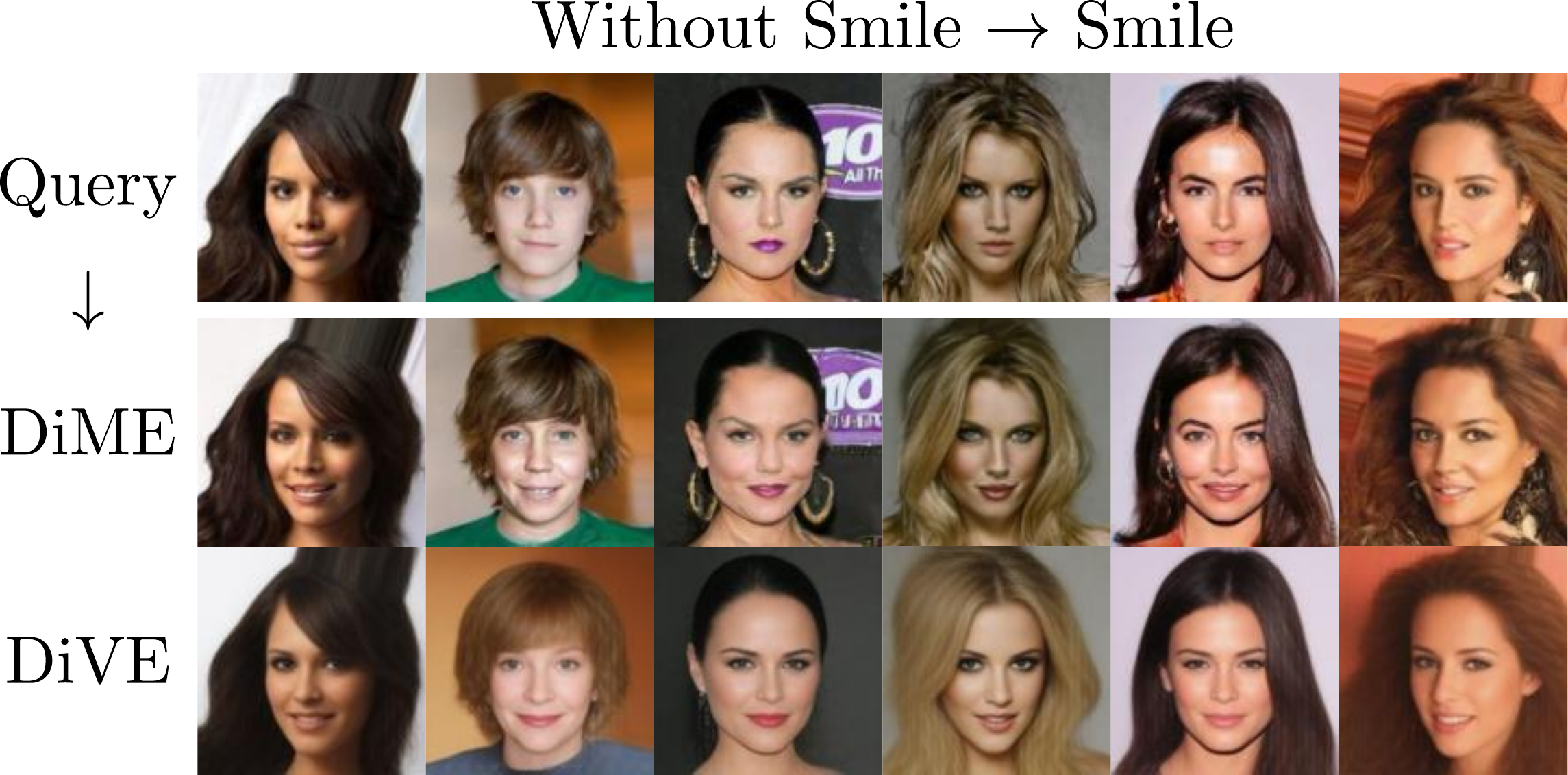}
    \caption{\textbf{Counterfactual explanations for the \textit{Smile}.} We visualize DiME and DiVE explanations targeting the label No Bags under the eyes.}
    \label{fig:ns-t-s}
\end{figure}

\begin{figure}
    \centering
    \includegraphics[width=0.9\textwidth]{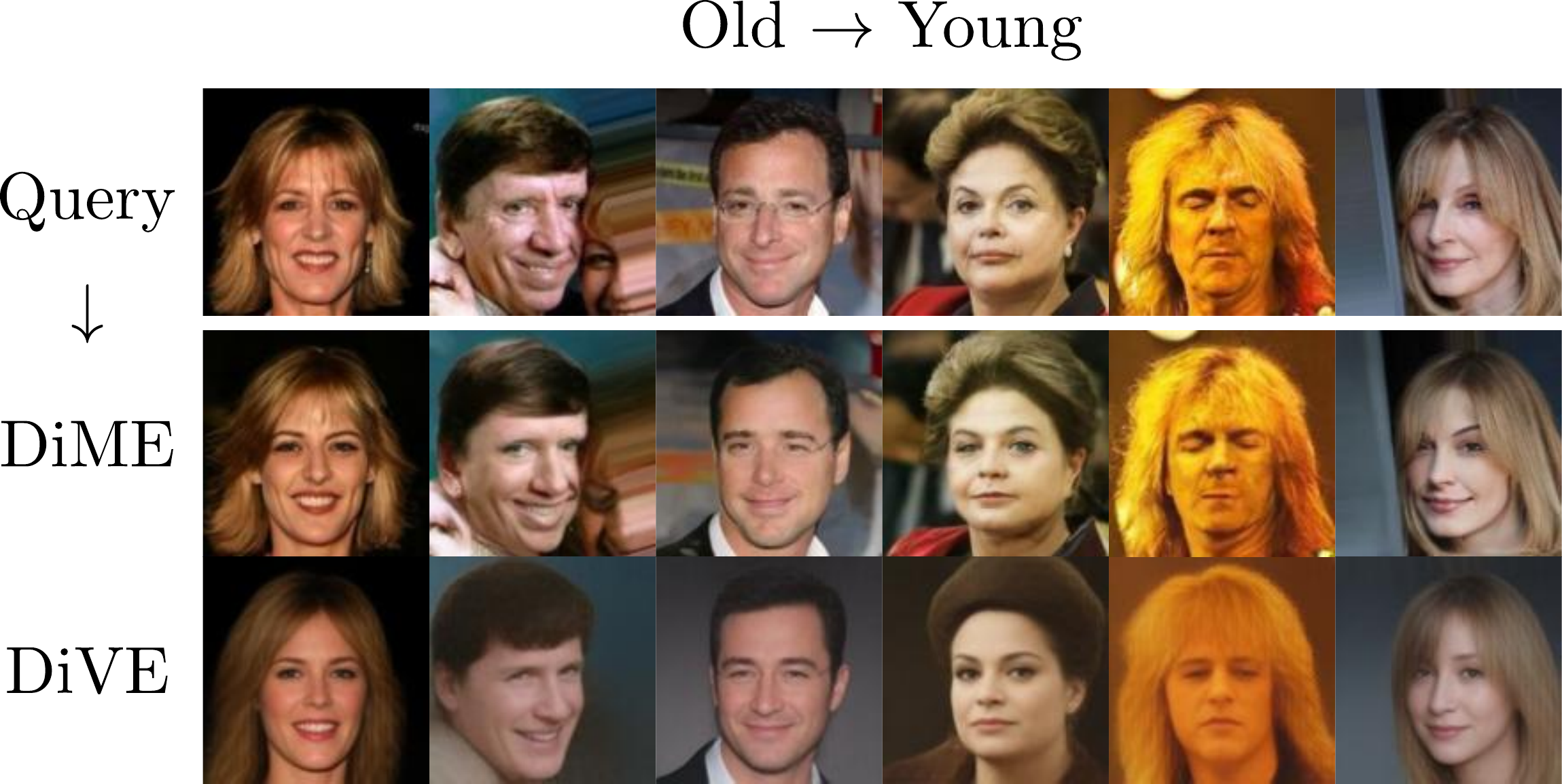}
    \caption{\textbf{Counterfactual explanations for the \textit{Age} attribute.} We visualize DiME and DiVE explanations targeting the label Young.}
    \label{fig:o-t-y}
\end{figure}

\begin{figure}
    \centering
    \includegraphics[width=0.9\textwidth]{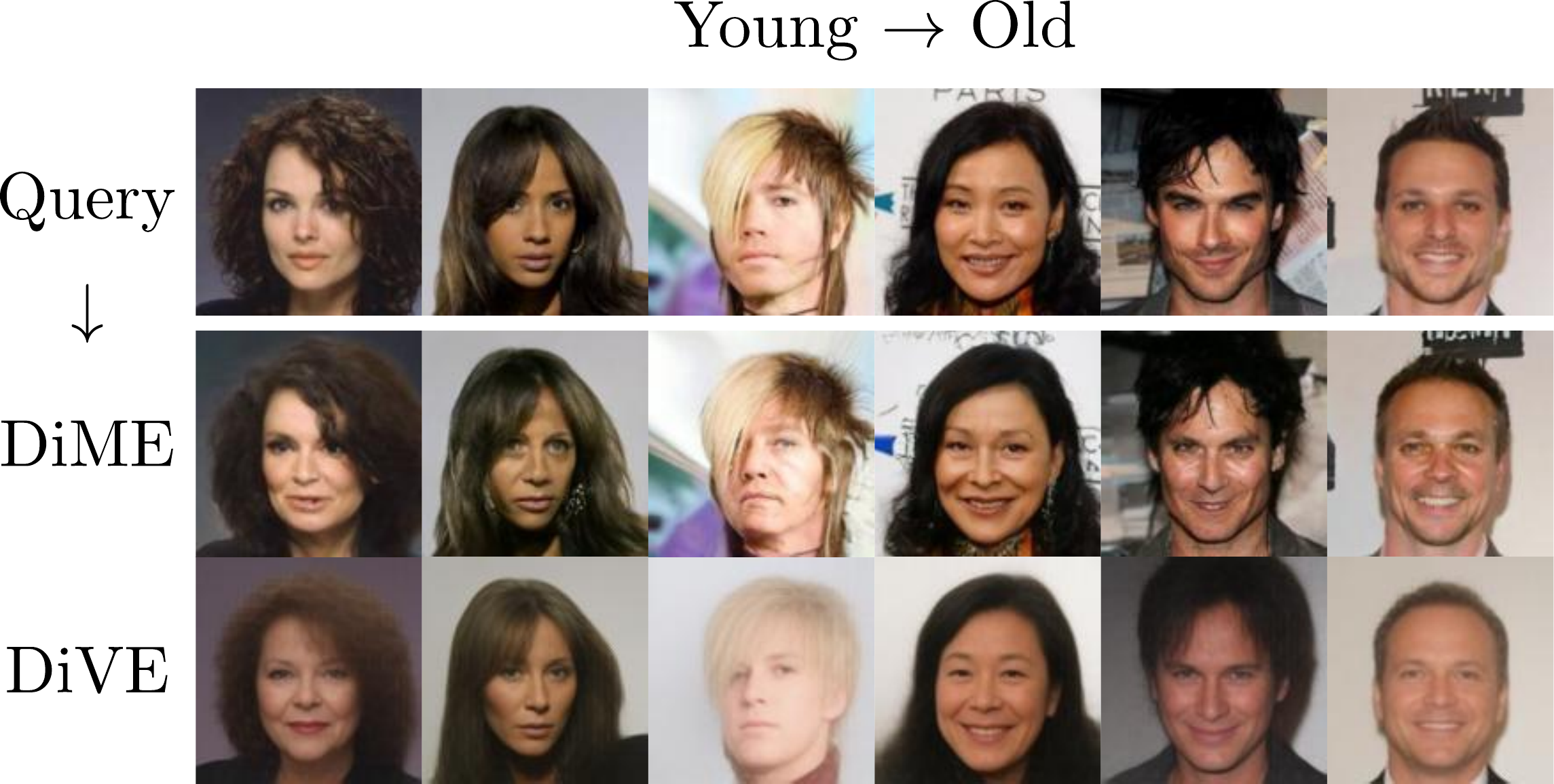}
    \caption{\textbf{Counterfactual explanations for the \textit{Age} attribute.} We visualize DiME and DiVE explanations targeting the label Old.}
    \label{fig:y-t-o}
\end{figure}

\begin{figure}[h]
    \centering
    \includegraphics[width=0.9\textwidth]{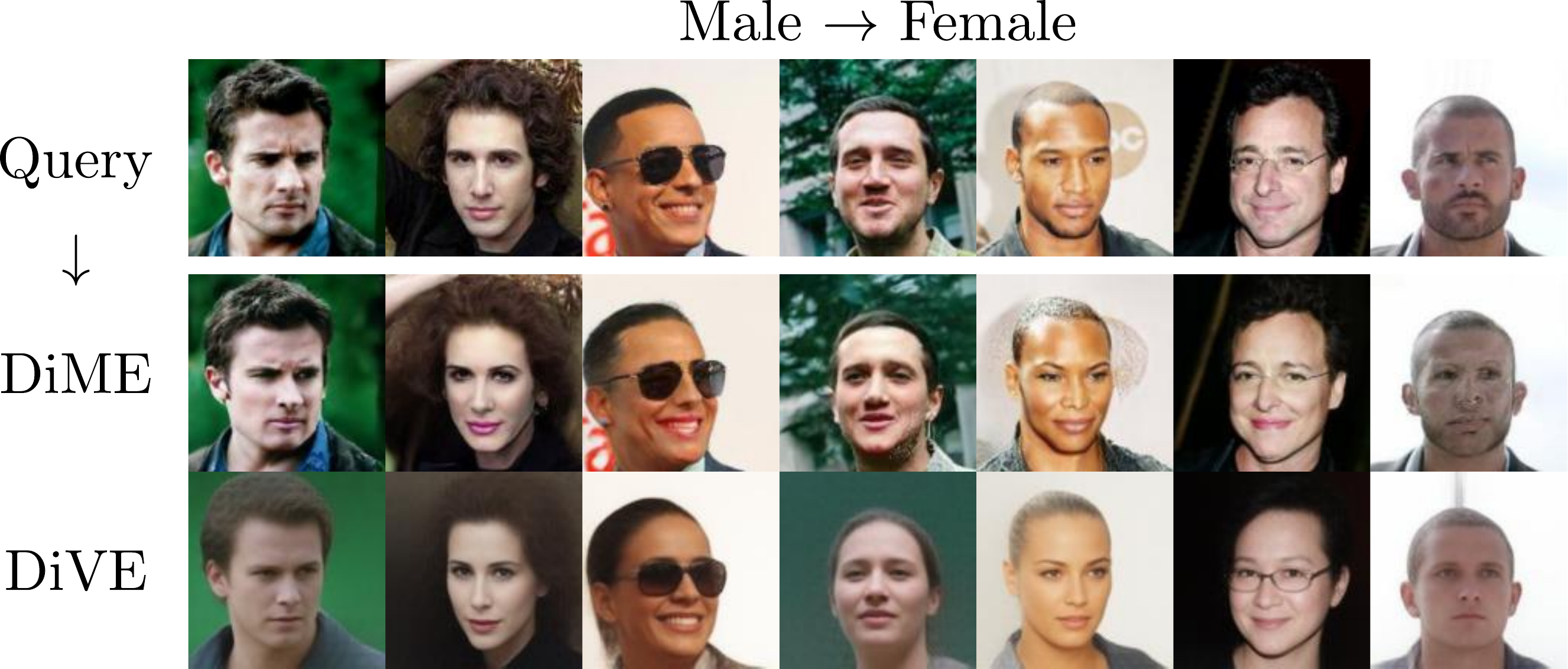}
    \caption{\textbf{Counterfactual explanations from \textit{male} to \textit{female}.} We visualize DiME and DiVE explanations.}
    \label{fig:male-to-female}
\end{figure}

\begin{figure}[h]
    \centering
    \includegraphics[width=0.9\textwidth]{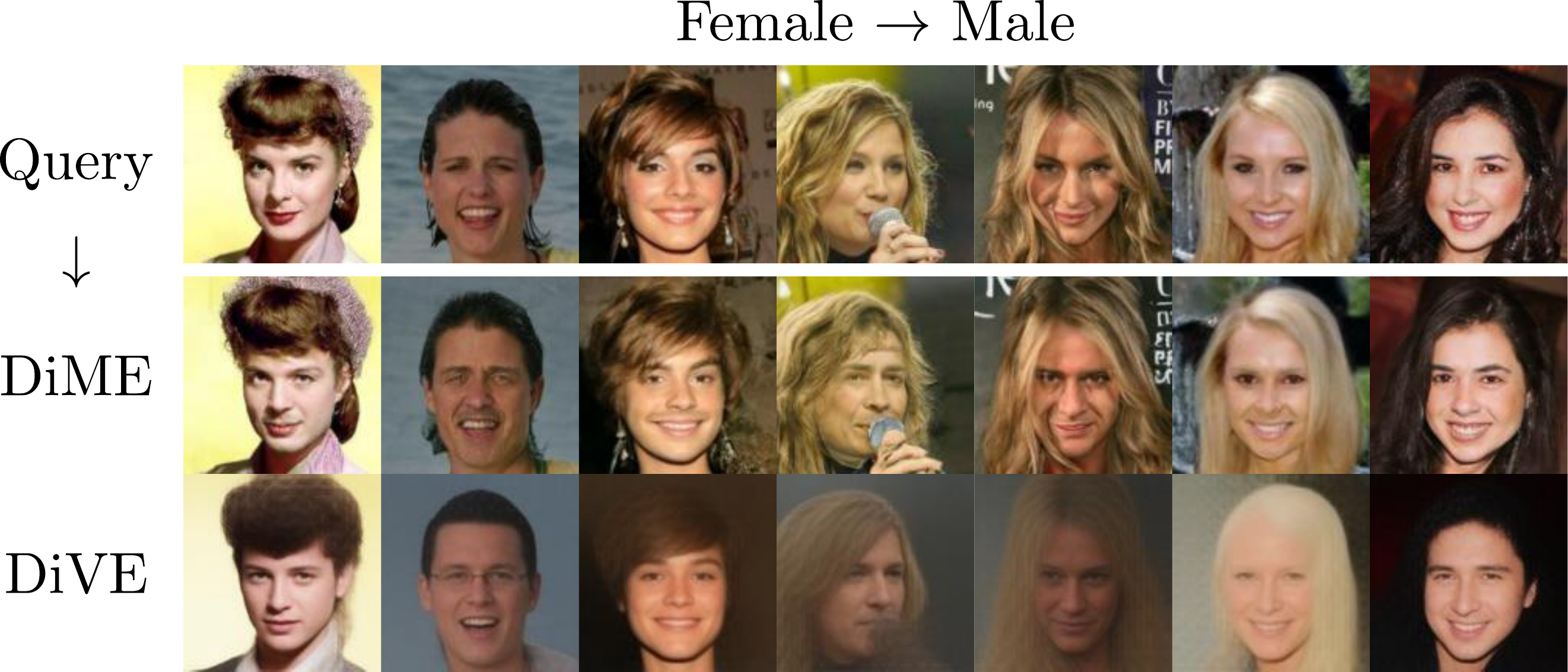}
    \caption{\textbf{Counterfactual explanations from \textit{female} to \textit{male}.} We visualize DiME and DiVE explanations.}
    \label{fig:female-to-male}
\end{figure}

\begin{figure}[h]
    \centering
    \includegraphics[width=0.90\textwidth]{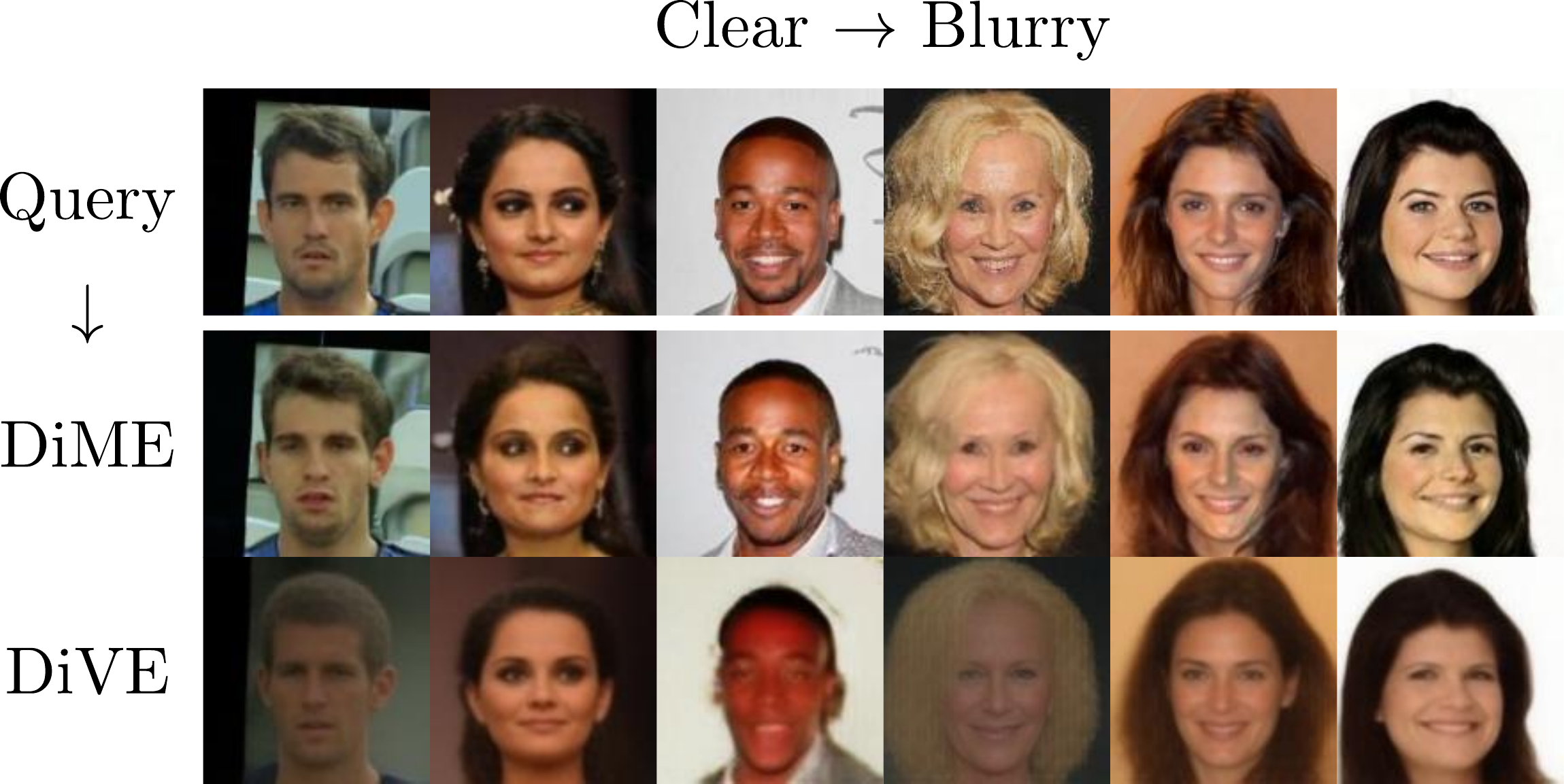}
    \caption{\textbf{Counterfactual explanations for the attribute \textit{Blurry}.} We visualize DiME and DiVE explanations using as target to deblur the input instance.}
    \label{fig:blurry-to-clear}
    \vspace{-0.4cm}
\end{figure}

\begin{figure}[h]
    \centering
    \includegraphics[width=0.90\textwidth]{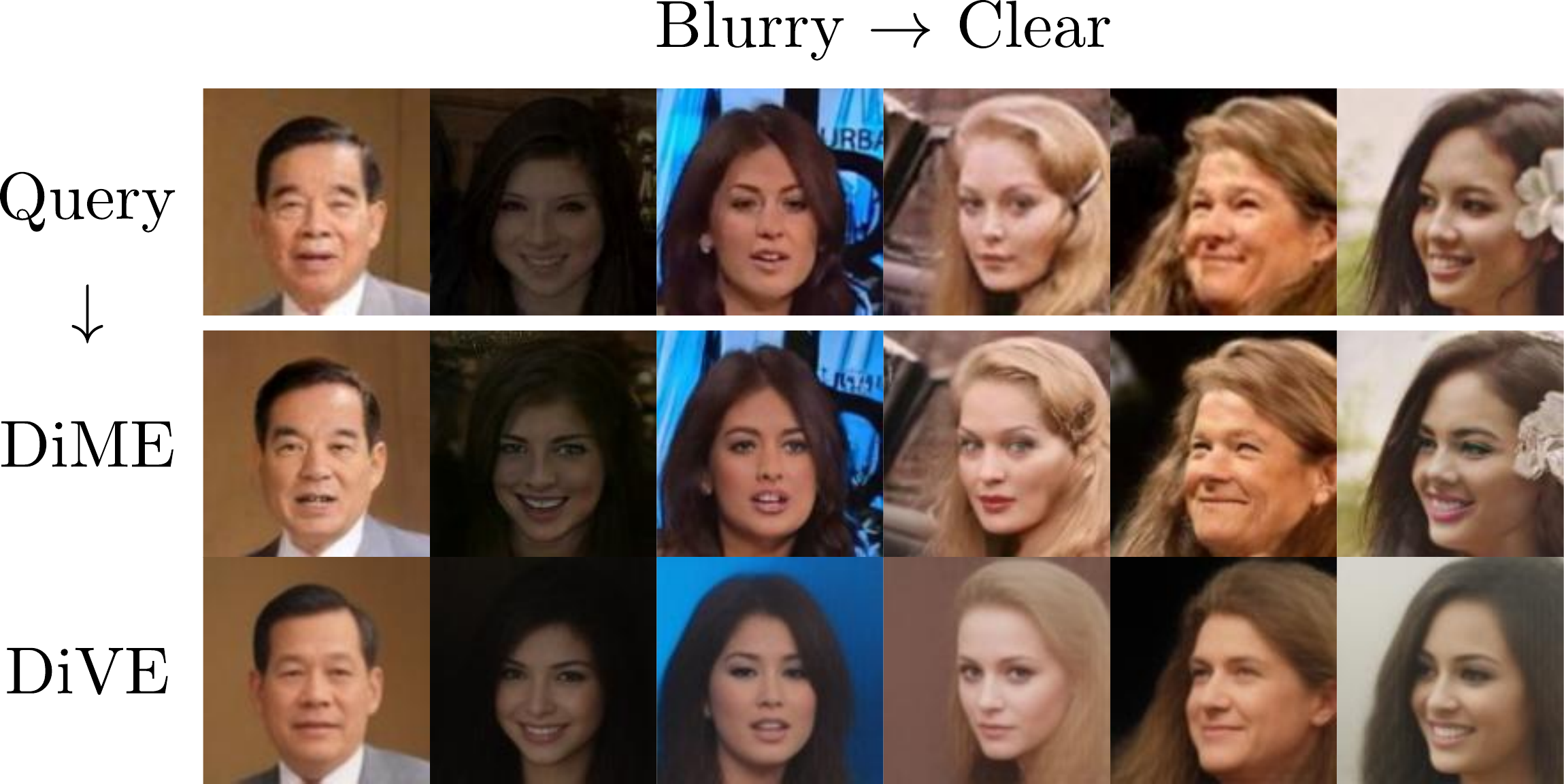}
    \caption{\textbf{Counterfactual explanations for the attribute \textit{Blurry}.} We visualize DiME and DiVE explanations using as target the blurry.}
    \label{fig:clear-to-blurry}
\end{figure}

\begin{figure}
    \centering
    \includegraphics[width=0.9\textwidth]{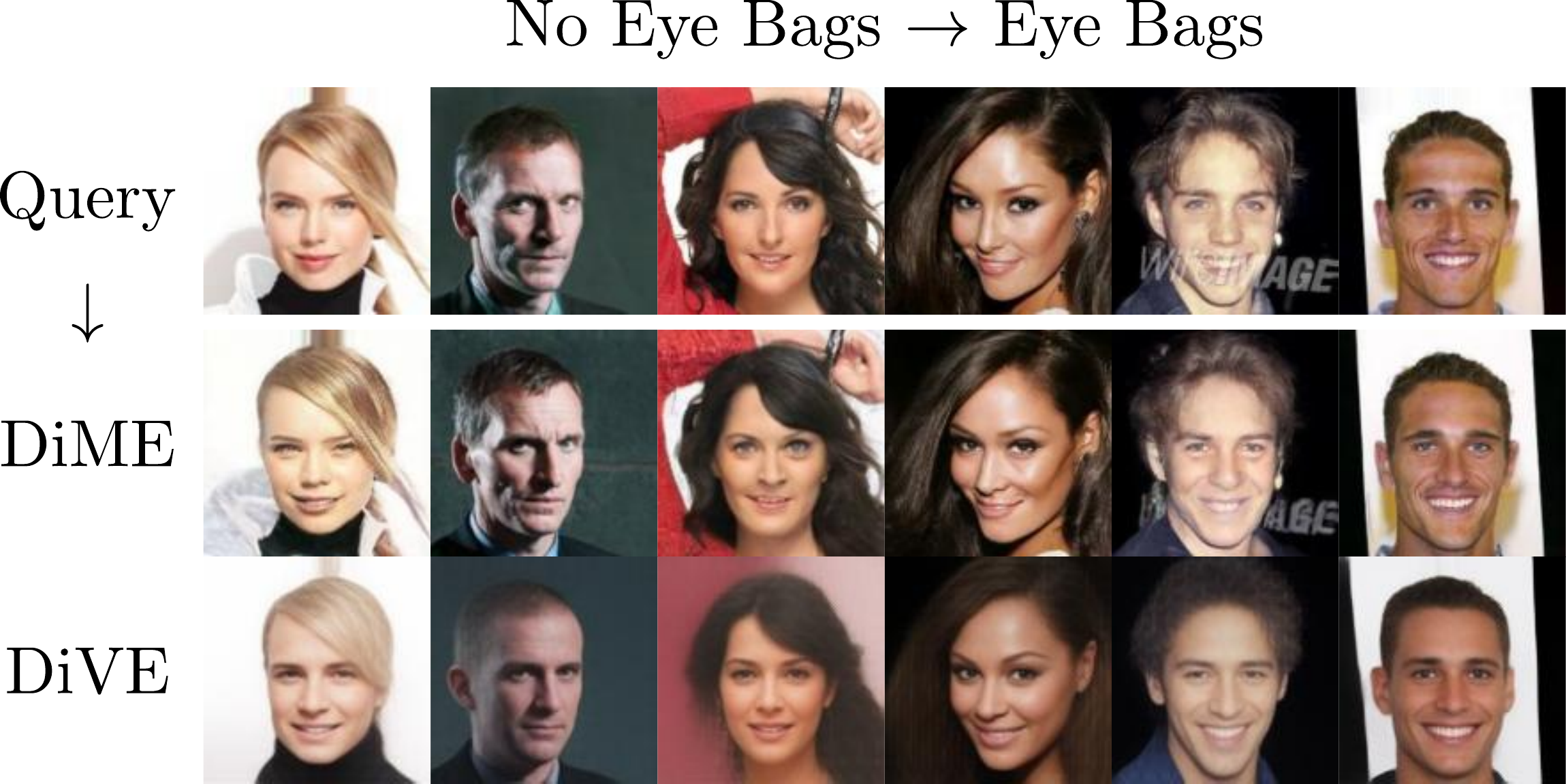}
    \caption{\textbf{Counterfactual explanations for the \textit{Bag Under Eye}.} We visualize DiME and DiVE explanations targeting the label Bags under the eyes.}
    \label{fig:nobag-to-bags}
    \vspace{-0.4cm}
\end{figure}

\begin{figure}
    \centering
    \includegraphics[width=0.9\textwidth]{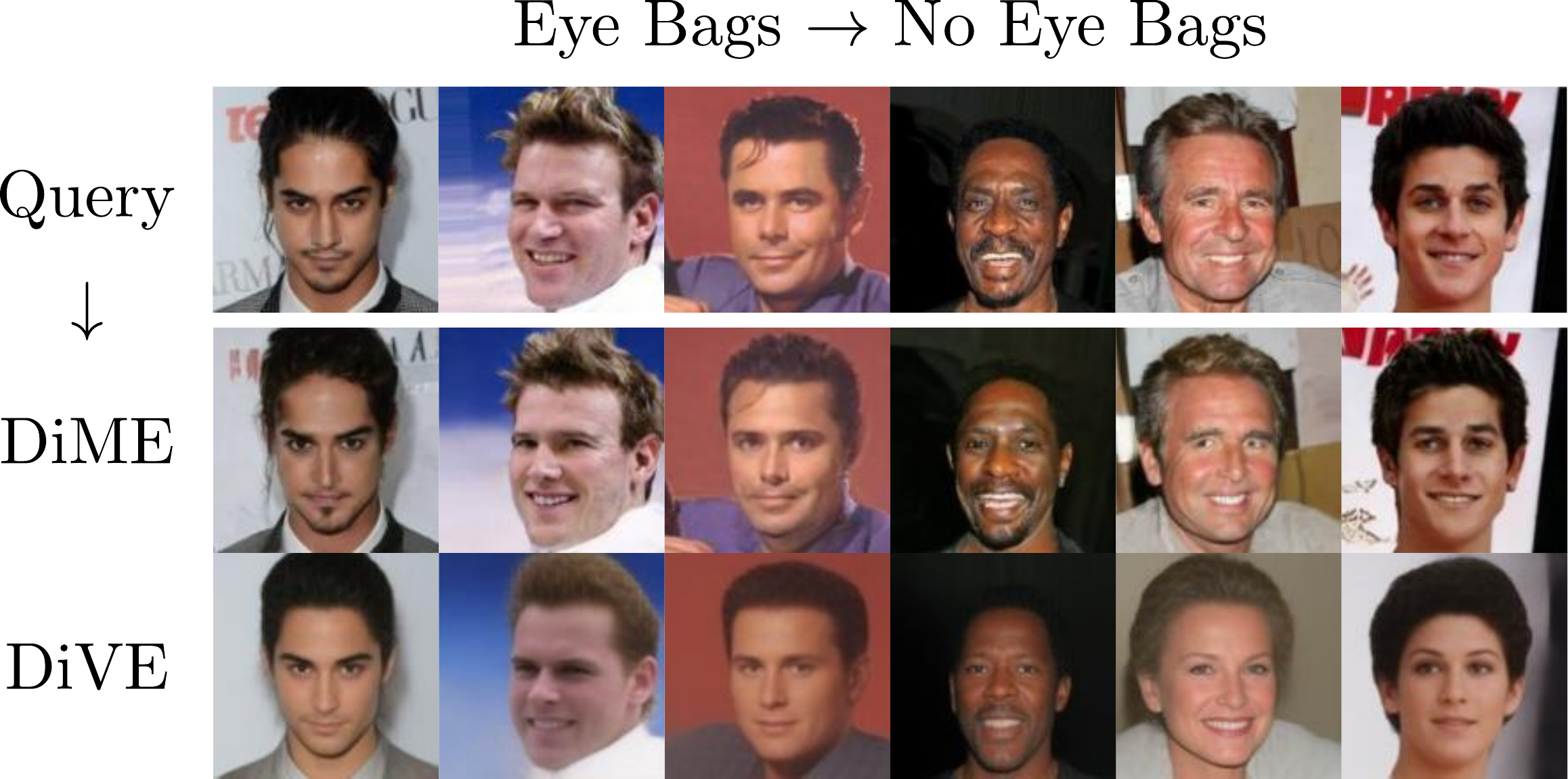}
    \caption{\textbf{Counterfactual explanations for the \textit{Bag Under Eye}.} We visualize DiME and DiVE explanations targeting the label No Bags under the eyes.}
    \label{fig:bags-to-nobags}
\end{figure}

\begin{figure}[!h]
    \centering
    \begin{subfigure}[b]{1\textwidth}
        \centering
        \includegraphics[width=0.9\textwidth]{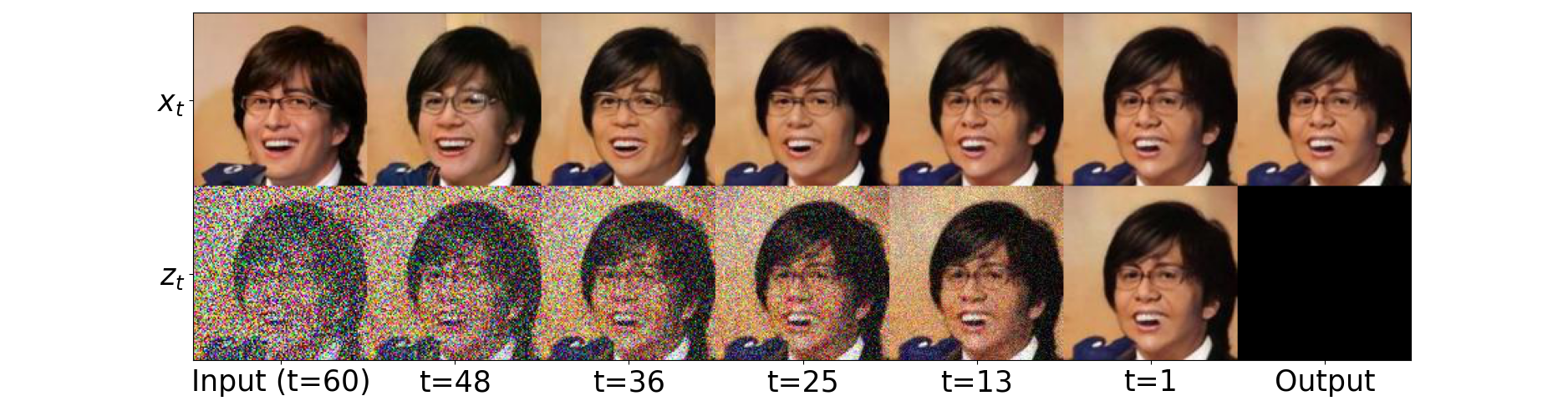}
    \end{subfigure}
    \begin{subfigure}[b]{1\textwidth}
        \centering
        \includegraphics[width=0.9\textwidth]{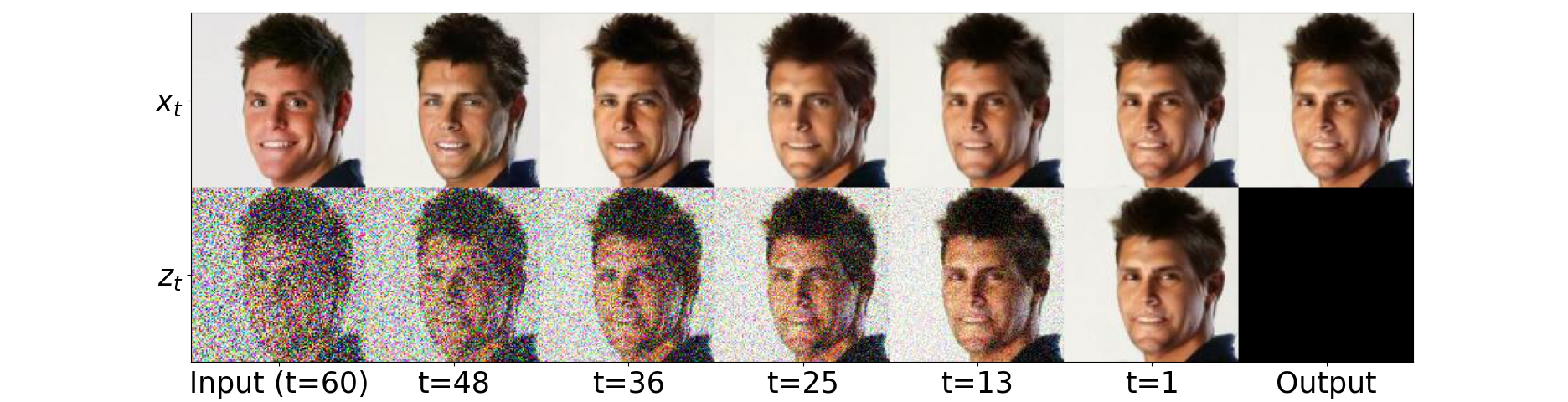}
    \end{subfigure}
    \begin{subfigure}[b]{1\textwidth}
        \centering
        \includegraphics[width=0.9\textwidth]{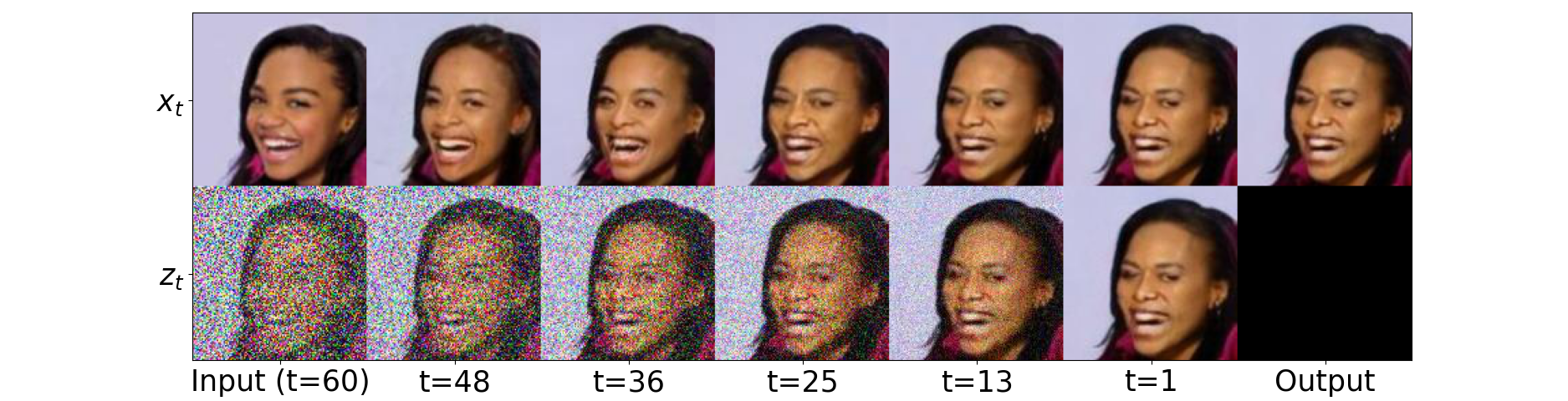}
    \end{subfigure}
    \begin{subfigure}[b]{1\textwidth}
        \centering
        \includegraphics[width=0.9\textwidth]{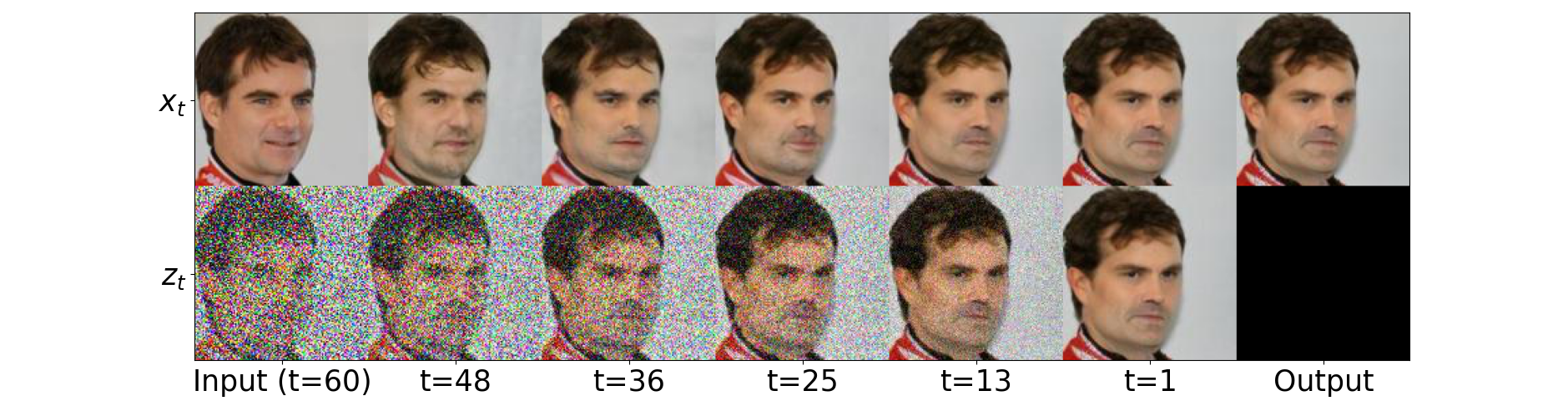}
    \end{subfigure}
    \begin{subfigure}[b]{1\textwidth}
        \centering
        \includegraphics[width=0.9\textwidth]{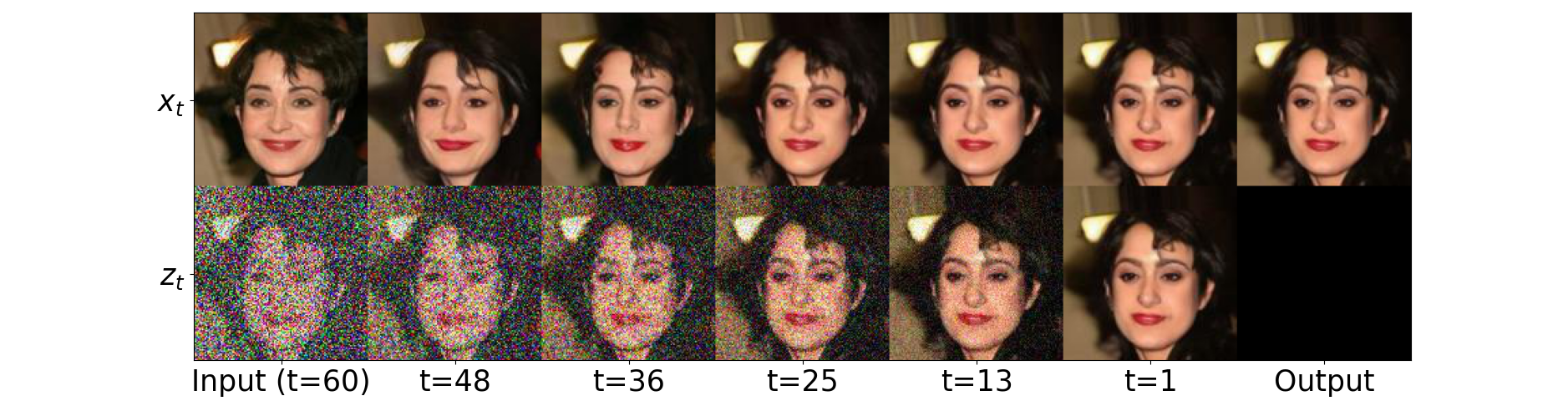}
    \end{subfigure}
    \caption{\textbf{Visual inspection over $t$.} We visualize the evolution of the noisy $z_t$ and clean instances $x_t$ over time. }
    \label{fig:nsev1}
\end{figure}

\begin{figure}
    \centering
    \begin{subfigure}[b]{1\textwidth}
        \centering
        \includegraphics[width=0.9\textwidth]{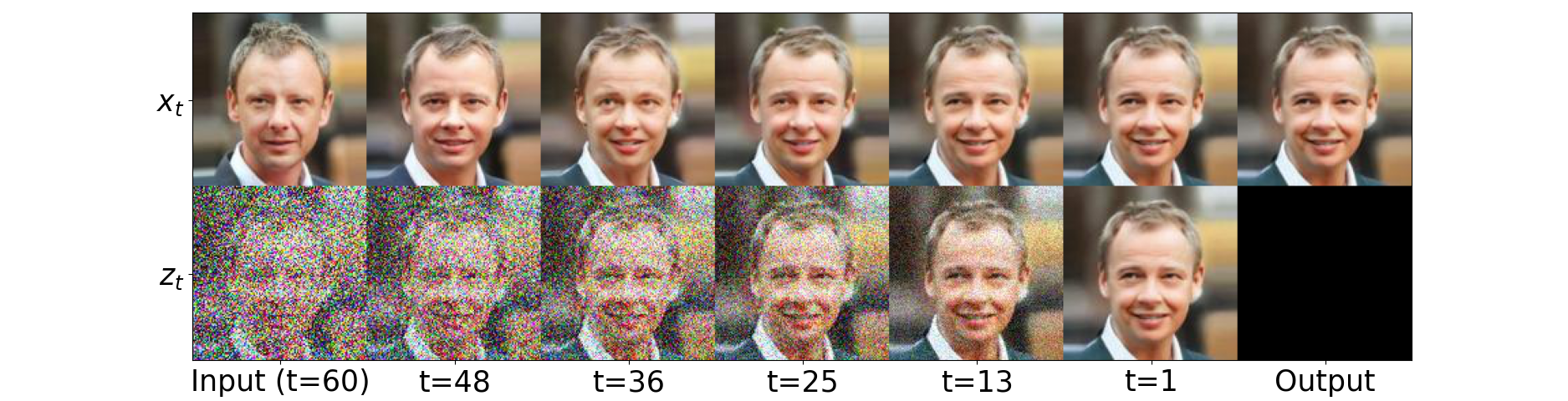}
    \end{subfigure}
    \begin{subfigure}[b]{1\textwidth}
        \centering
        \includegraphics[width=0.9\textwidth]{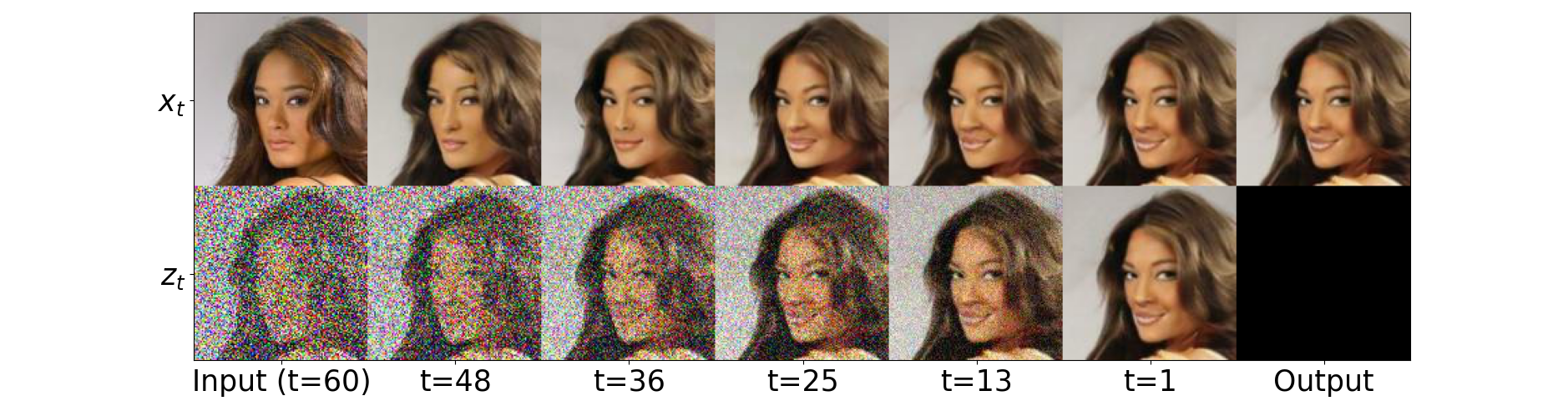}
    \end{subfigure}
    \begin{subfigure}[b]{1\textwidth}
        \centering
        \includegraphics[width=0.9\textwidth]{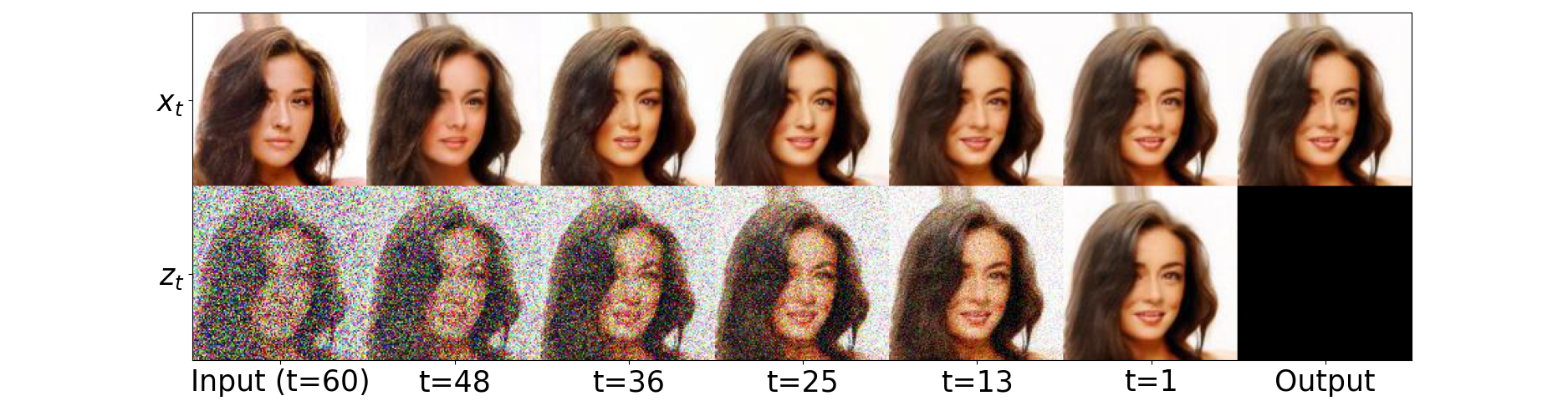}
    \end{subfigure}
    \begin{subfigure}[b]{1\textwidth}
        \centering
        \includegraphics[width=0.9\textwidth]{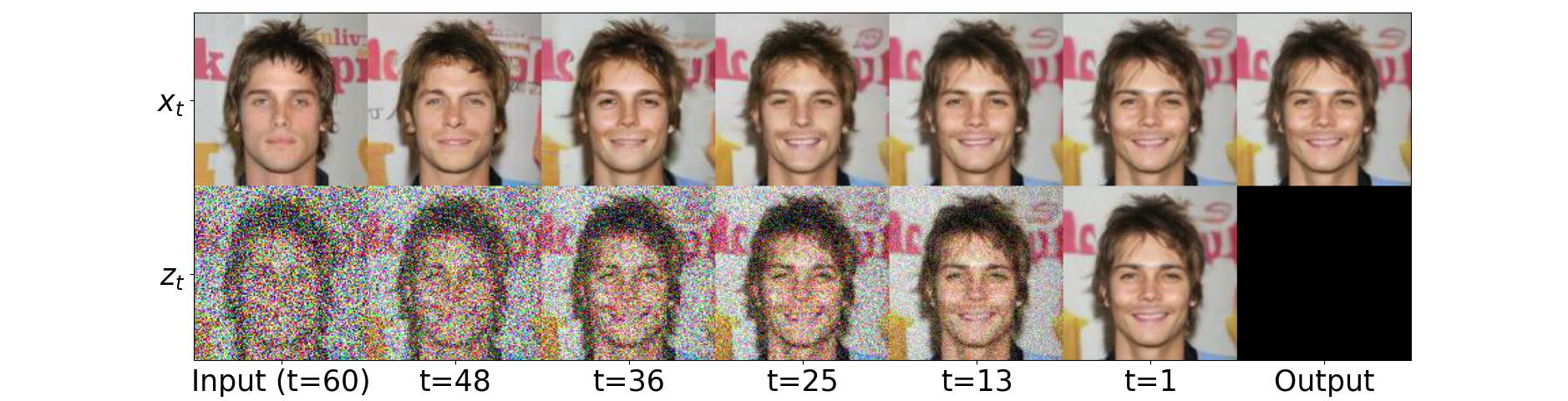}
    \end{subfigure}
    \begin{subfigure}[b]{1\textwidth}
        \centering
        \includegraphics[width=0.9\textwidth]{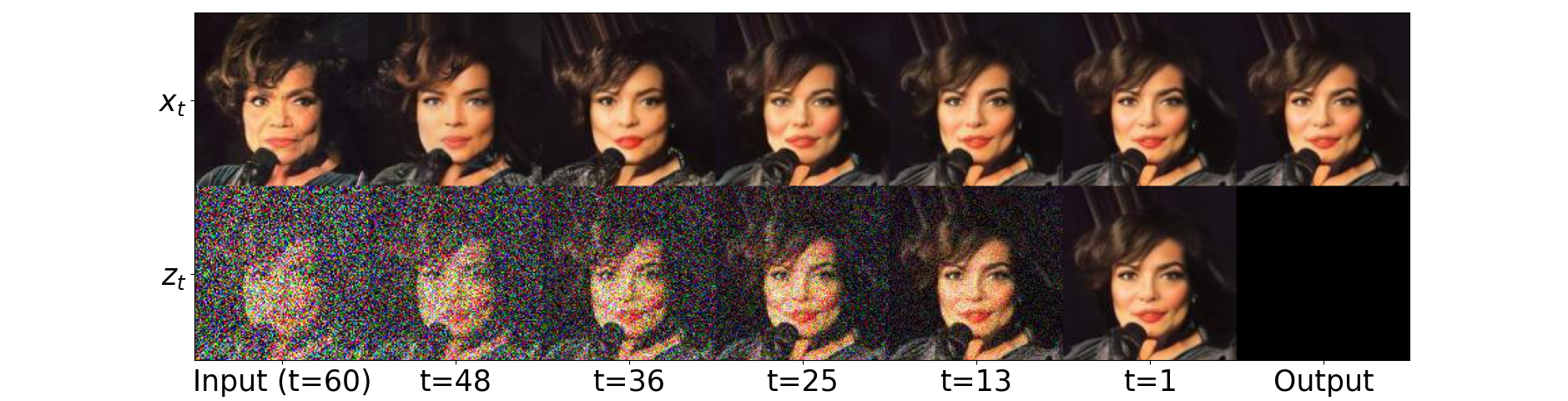}
    \end{subfigure}
    \caption{\textbf{Visual inspection over $t$.} We visualize the evolution of the noisy $z_t$ and clean instances $x_t$ over time. }
    \label{fig:sev1}
\end{figure}

\begin{figure}
    \centering
    \begin{subfigure}[b]{1\textwidth}
        \includegraphics[width=\textwidth]{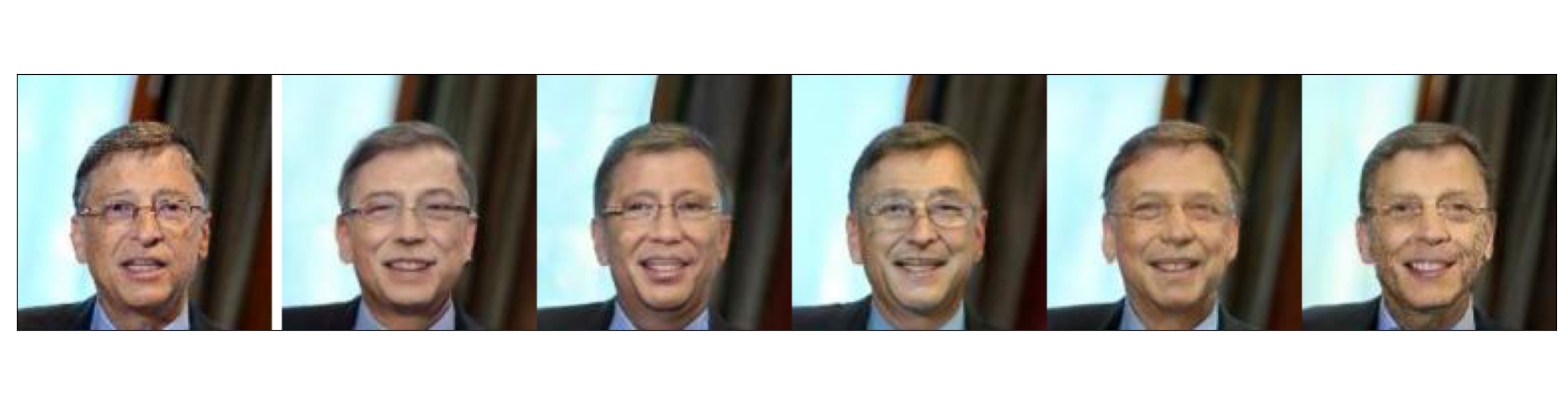}
    \end{subfigure}
    \begin{subfigure}[b]{1\textwidth}
        \includegraphics[width=\textwidth]{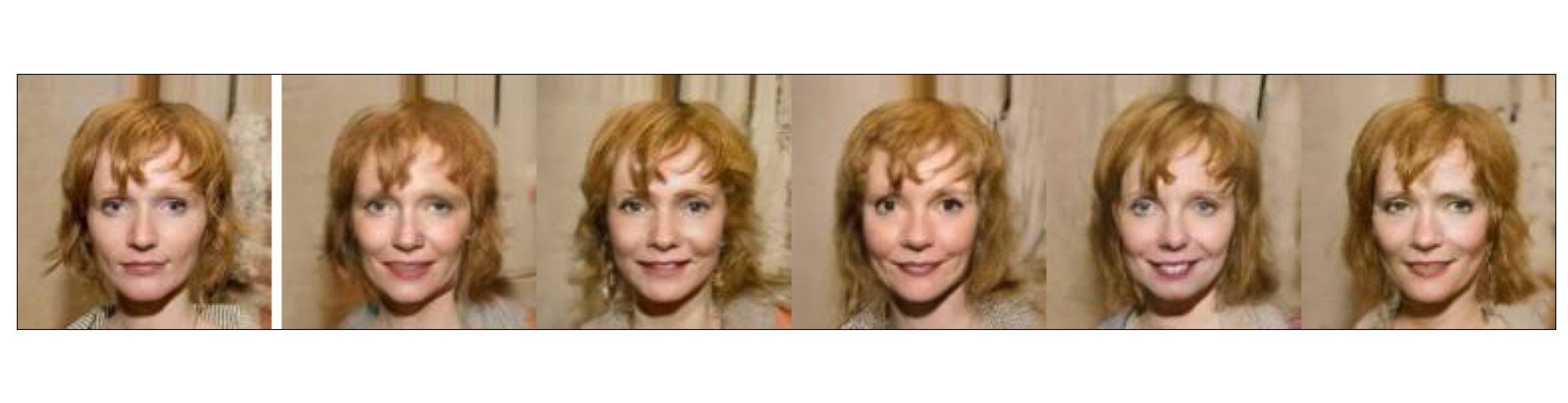}
    \end{subfigure}
    \begin{subfigure}[b]{1\textwidth}
        \includegraphics[width=\textwidth]{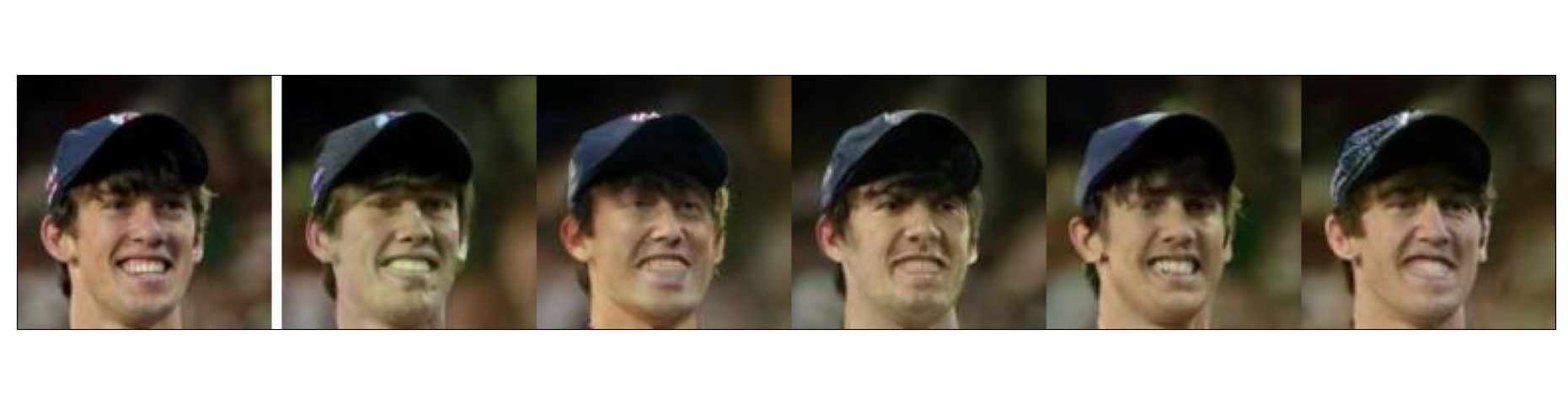}
    \end{subfigure}
    \begin{subfigure}[b]{1\textwidth}
        \includegraphics[width=\textwidth]{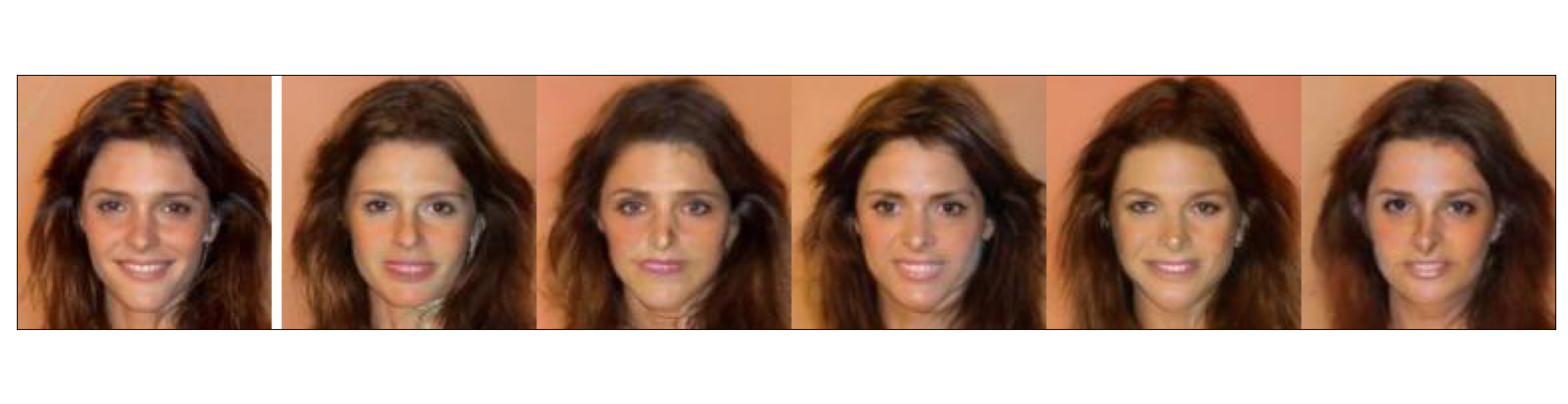}
    \end{subfigure}
    
    \caption{\textbf{Variability examples.} We visualize the effects of the stochasticity of DiME to produce multiple explanations.}
    \label{fig:variability-sup1}
\end{figure}

\begin{figure}
    \centering
    \begin{subfigure}[b]{1\textwidth}
        \includegraphics[width=\textwidth]{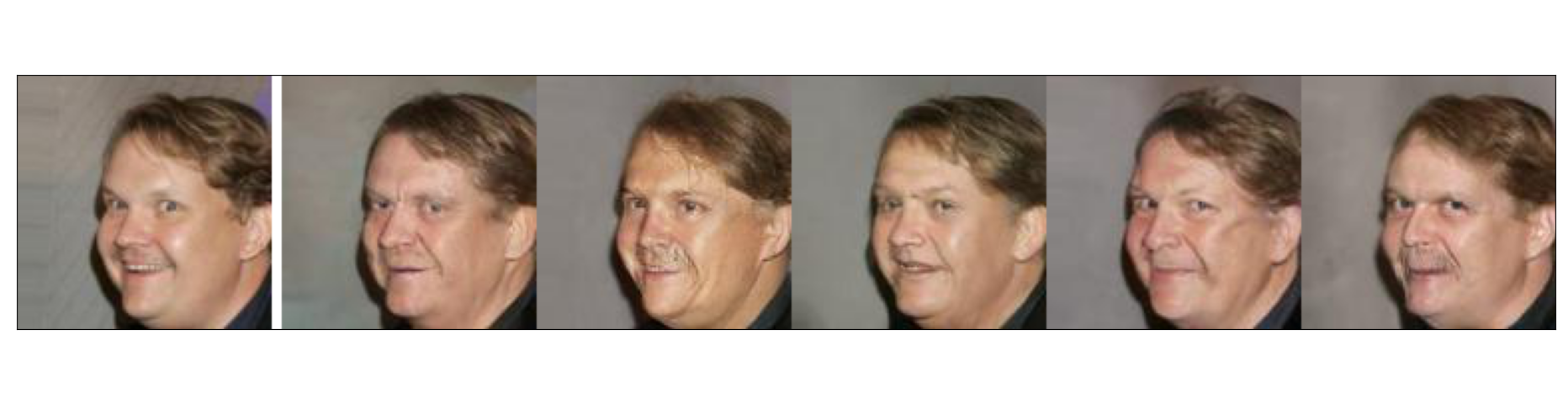}
    \end{subfigure}
    \begin{subfigure}[b]{1\textwidth}
        \includegraphics[width=\textwidth]{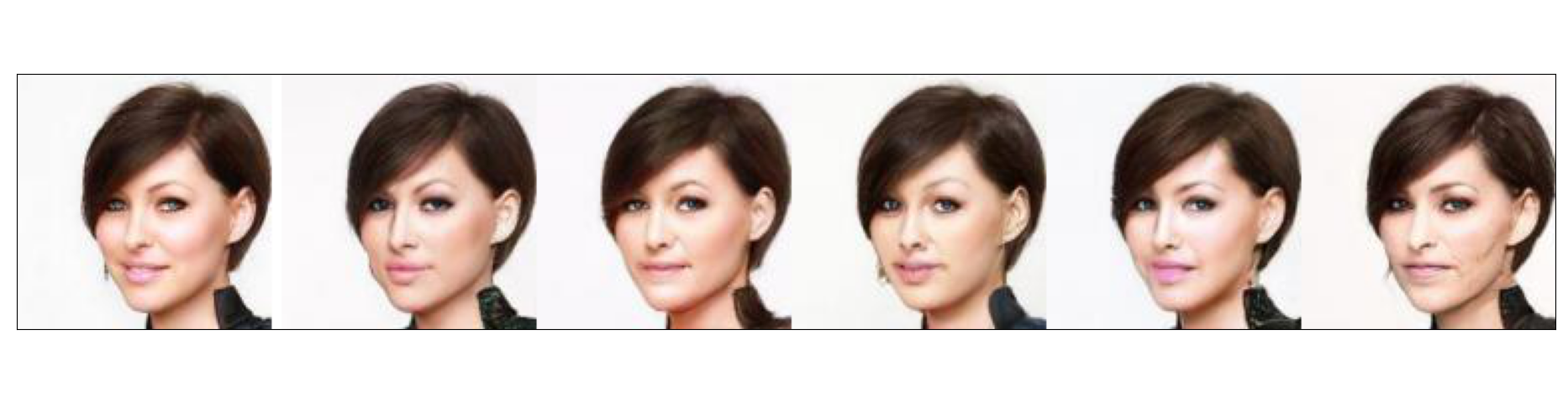}
    \end{subfigure}
    \begin{subfigure}[b]{1\textwidth}
        \includegraphics[width=\textwidth]{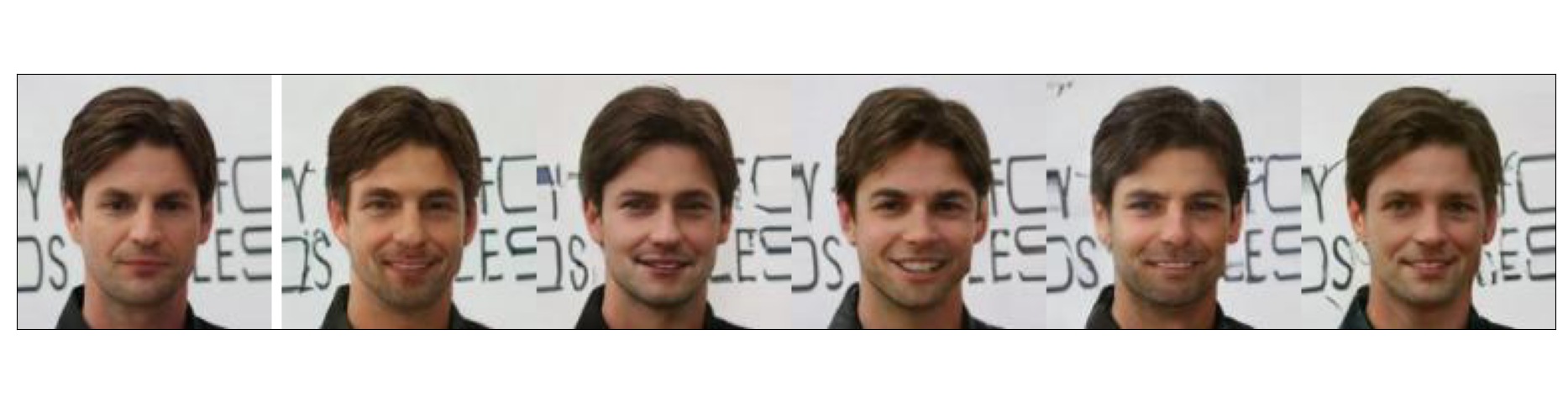}
    \end{subfigure}
    \begin{subfigure}[b]{1\textwidth}
        \includegraphics[width=\textwidth]{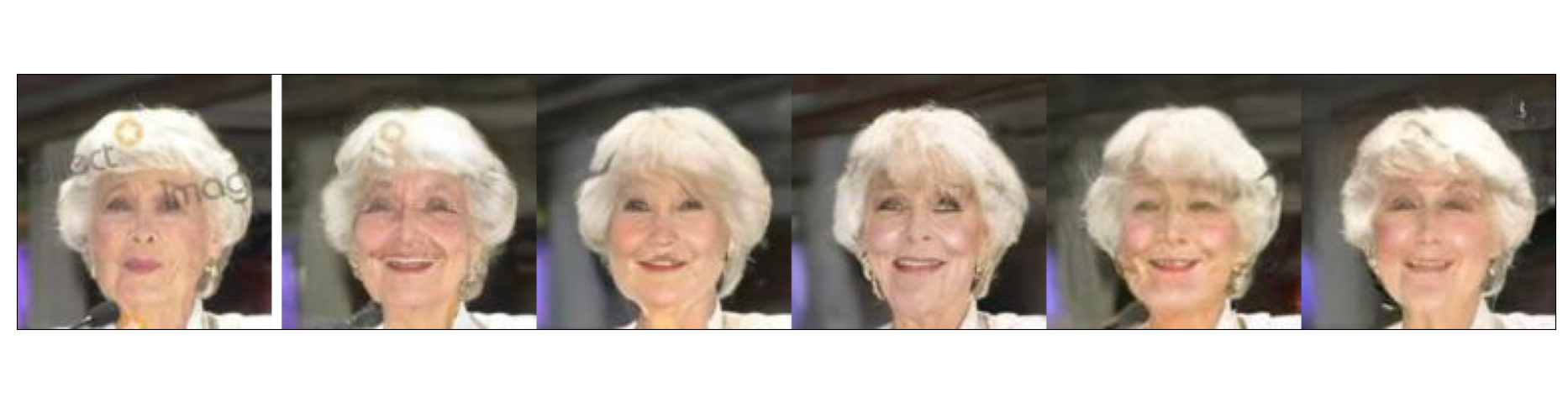}
    \end{subfigure}
    
    \caption{\textbf{Variability examples.} We visualize the effects of the stochasticity of DiME to produce multiple explanations.}
    \label{fig:variability-sup2}
\end{figure}

In this section, we visualize some qualitative results from our proposed benchmark for counterfactual explanations. We include cases for the smile, young and other attributes from the CelebA dataset. 
Also, we compare our results with DiVE's explanations in Figures~\ref{fig:s-t-ns} to~\ref{fig:bags-to-nobags}.
Further, we show some examples of the evolution at each time step of the noisy and clean instances in Figures~\ref{fig:nsev1} and~\ref{fig:sev1}. Finally, we visualize more examples on the variability of DiME in Figures~\ref{fig:variability-sup1} and~\ref{fig:variability-sup2}.

In general, we see a clear pattern comparing DiME and DiVE. DiME's generated instances are closer to the query image than DiVE's. Further, DiVE uses a VAE as the generative model, so their CEs are blurrier than ours. 

\clearpage

With respect to the gender attribute, we visualize two differences between each gender. 
For the \textit{male to female} case, DiME exposes a clear correlation between the female label and the attributes \textit{heavy makeup} and \textit{lipstick}. 
We suspect that the classifier mainly relies on these attributes to classify an image as a woman. 
In contrast, DiVE adds ``women-like'' features to flip the prediction. 
For the \textit{female to male} counterfactuals, major changes in the image are done to add female qualities for both models. 
The last two examples show that removing the makeup is enough the flip the classifier prediction.

\clearpage

Regarding the blurry attribute, at first glance, we see that DiVE's VAE helps blur the input instance. 
Nevertheless, as we see in Fig.~\ref{fig:blurry-to-clear}, CelebA's inherited blurry attribute is different from the one produced by DiVE.

\clearpage

The attribute \textit{Bags under the Eyes} has a clear and punctual location in the image: the region below the eyes. 
Both algorithms provide successful explanations when targeting this attribute. 
The main difference between DiME and DiVE performances is the capacity of DiME to retain as much fine-grained information as possible such as the hair, hands, and the background.

\clearpage

Following the study on the evolution of the clean images $x_t$ on time, we display more examples along with their noisy version. 
We see that, when $t=48$ and $t=36$, the clean images present the most changes, while the last images do not vary much.

\clearpage

The following Figures visualize more examples of DiME's capacity to create diverse counterfactual explanations. 
The visualizations show that DiME retains most details when generating counterfactuals.

\end{document}